\documentclass[11pt,3p]{elsarticle}
\makeatletter
\def\ps@pprintTitle{\let\@oddhead\@empty
  \let\@evenhead\@empty
  \def\@oddfoot{\reset@font\hfil\thepage\hfil}
  \let\@evenfoot\@oddfoot
}
\makeatother

\usepackage{amsmath}
\usepackage{amsfonts}
\usepackage{amssymb}
\usepackage{bbm}
\usepackage{amsthm}
\usepackage{stmaryrd}

\usepackage{amscd}
\usepackage{mathtools}
\usepackage{mathrsfs}
\usepackage{commath}
\usepackage{lmodern}
\usepackage{color}
\usepackage{a4wide}
\usepackage{bm} 
\usepackage{physics}
\usepackage{enumitem}  
\usepackage[normalem]{ulem}
\usepackage{hyperref}
\hypersetup{ colorlinks=true, urlcolor  = blue, linkcolor = blue,
citecolor = blue}
\usepackage{booktabs}
\usepackage[export]{adjustbox}

\usepackage[english]{babel}
\usepackage[IL2]{fontenc}
\usepackage[utf8]{inputenc}
\usepackage{comment}
\usepackage{svg}
\usepackage[linesnumbered,ruled,vlined]{algorithm2e}
\usepackage[noend]{algpseudocode}
\SetKwInput{KwInput}{Input}                
\SetKwInput{KwOutput}{Output}              

\SetCommentSty{mycommfont}

\usepackage{calrsfs}
\DeclareMathAlphabet{\pazocal}{OMS}{zplm}{m}{n}
\newcommand{\Lb}{\pazocal{L}}

\theoremstyle{definition}

\theoremstyle{remark}

\newcommand{\marius}[1]{\textcolor{black}{ #1} }

\begin{document}

\begin{frontmatter}

\title{Self-induced stochastic resonance: A physics-informed machine learning approach}
\author[1]{Divyesh Savaliya}
\author[1]{Marius E. Yamakou}
\ead{marius.yamakou@fau.de}
\address[1]{Department of Data Science, Friedrich-Alexander-Universit\"at Erlangen-N\"urnberg, N\"urnberger Str. 74, 91052 Erlangen, Germany}

\begin{abstract}
Self-induced stochastic resonance (SISR) is the emergence of coherent oscillations in slow-fast excitable systems driven solely by noise, without external periodic forcing or proximity to a bifurcation. This work presents a physics-informed machine learning framework for modeling and predicting SISR in the stochastic FitzHugh–Nagumo neuron. We embed the governing stochastic differential equations and SISR-asymptotic timescale-matching constraints directly into a Physics-Informed Neural Network (PINN) based on a Noise-Augmented State Predictor architecture. The composite loss integrates data fidelity, dynamical residuals, and barrier-based physical constraints derived from Kramers’ escape theory. 
The trained PINN accurately predicts the dependence of spike-train coherence on noise intensity, excitability, and timescale separation, matching results from direct stochastic simulations with substantial improvements in accuracy and generalization compared with purely data-driven methods, requiring significantly less computation. The framework provides a data-efficient and interpretable surrogate model for simulating and analyzing noise-induced coherence in multiscale stochastic systems.
\end{abstract}

\end{frontmatter}

\section{Introduction}
\label{sec:introduction}
Noise, often regarded as a source of disorder and unpredictability, can in fact play a constructive role in nonlinear dynamical systems~\cite{freidlin2001stochastic,berglund2006noise,roy2022role}. In a variety of physical, biological, and engineering contexts, random fluctuations can induce or enhance regularity in systems that are otherwise quiescent or weakly active. This counterintuitive phenomenon—where noise promotes order rather than destroys it—has been extensively studied under the umbrella of stochastic resonance and its generalizations \cite{hanggi2002stochastic,gluckman1998stochastic,bulsara1991stochastic}. In excitable systems such as neurons, noise can trigger coherent oscillations that would not occur in the absence of random fluctuations, thereby shaping information transmission and signal reliability in the nervous system \cite{guoFunctionalImportanceNoise2018}.

Among the diverse manifestations of noise-induced coherence, self-induced stochastic resonance (SISR) \cite{muratovSelfinducedStochasticResonance2005,yamakouCoherentNeuralOscillations2018,yamakouControlCoherenceResonance2019} occupies a particularly distinctive position. Classical stochastic resonance (SR) requires the simultaneous presence of noise and a weak periodic forcing~\cite{hanggi2002stochastic,longtin1993stochastic}, with the noise optimally amplifying the external signal. Coherence resonance (CR) \cite{pikovsky1997coherence,yamakou2023combined,deville2005two}, in contrast, arises in autonomous systems tuned close to a bifurcation threshold \cite{jiaCoherenceresonanceinducedNeuronalFiring2011}, where noise enhances the intrinsic tendency toward oscillations without any external drive. SISR differs fundamentally from both: it occurs in excitable systems whose deterministic dynamics possess only stable fixed point(s), far from any bifurcation \cite{muratovSelfinducedStochasticResonance2005,deville2005two}, no external periodic forcing is required, and, unlike SR and CR, it requires a strong timescale separation between the system variables. In this regime, no deterministic oscillations exist, yet weak stochastic perturbations alone can generate highly regular, self-sustained oscillations. 

The mechanism behind the SISR phenomenon lies in the matching between two characteristic timescales—the deterministic relaxation along the slow manifold and the stochastic escape time across potential barriers—first elucidated through Kramers’ escape-rate theory and large deviation analysis \cite{muratovSelfinducedStochasticResonance2005,yamakouCoherentNeuralOscillations2018,zhu2021stochastic}. When these timescales coincide, noise-induced transitions occur at reproducible points in phase space, producing coherent rhythmic activity. Thus, SISR reveals how random fluctuations, rather than an external drive or proximity to instability, can organize slow–fast dynamics into emergent periodic behavior, bridging nonequilibrium statistical physics and neuronal excitability \cite{yamakouControlCoherenceResonance2019,zhu2024reduced,yamakouLevyNoiseinducedSelfinduced2022}.

Despite its conceptual significance, SISR remains analytically and numerically challenging to characterize. The phenomenon is inherently multiscale and stochastic: accurate prediction of coherence requires resolving both the slow deterministic drift of the recovery variable and the exponentially rare noise-induced escapes from potential wells. Classical numerical approaches, such as direct integration of the stochastic FitzHugh–Nagumo (FHN) equations, can reproduce SISR but at the cost of long simulations and extensive ensemble averaging to capture rare transitions \cite{zhu2021stochastic,zhu2023self,yamakouOptimalSelfinducedStochastic2020}. Analytical approaches, including asymptotic analysis \cite{berglund2006noise,freidlin2012random}, large-deviation theory \cite{freidlin2001stochastic,freidlin2001stable,kuehn2015multiple}, and Kramers’ law \cite{kramers1940brownian}, provide deep theoretical insight into escape dynamics and resonance conditions, but their validity is often restricted to idealized limits (\textit{e.g.,} weak noise, infinite timescale separation), limiting applicability to realistic parameter regimes.

Recent advances in scientific machine learning have opened new avenues for addressing such multiscale stochastic problems \cite{peng2020multiscale,arbabi2020linking}. In particular, physics-informed neural networks (PINNs) offer a promising framework by embedding \marius{the governing stochastic multiple} timescale equations and physical constraints directly into the learning process \cite{gao2024learning,raissiPhysicsInformedNeural2019,markidisOldNewPINNs2021}. By combining data-driven inference with dynamical priors derived from the system’s equations, PINNs can efficiently approximate solutions to stochastic differential equations even when only limited or noisy data are available. This hybrid approach enables not only the reconstruction of stochastic trajectories but also the identification of physically meaningful quantities such as potential barriers, escape rates, and timescale ratios that govern the emergence of SISR.

Beyond PINNs, there is a rapidly growing body of work on machine-learning 
surrogates for stochastic and multiscale dynamical systems. Deep generative 
models and neural surrogates have been proposed to emulate stochastic 
simulators and reproduce output distributions, enabling fast uncertainty 
quantification and rare-event statistics~\cite{thakur2022deep}. 
For multiscale stochastic systems with explicit slow--fast structure, data-driven methods have been developed to learn effective dynamics for the 
slow variables \cite{zielinski2022discovery}, discover low-dimensional embeddings, and construct reduced 
models that retain the essential stochastic behavior of the full system, 
including escape events and regime transitions~\cite{fabiani2024task,chen2024data}. 
In parallel, neural stochastic differential equations (neural SDEs) represent both drift and diffusion terms with neural networks and have emerged as flexible  surrogates for complex stochastic processes in continuous time, combining 
the expressive power of deep networks with SDE-based priors on temporal 
evolution~\cite{Tzen2019NeuralSDE,kidger2021efficient}. Our NASP--PINN 
framework is complementary to these approaches: it focuses on excitable 
slow--fast neuron dynamics, explicitly encodes the SISR timescale-matching 
conditions \marius{via the barrier-based constraints}, and leverages physics-informed 
training to obtain a noise-aware surrogate that faithfully reproduces coherent 
spiking induced solely by stochastic forcing.

It is worth mentioning that the prediction of coherence resonance, a distinct noise-induced resonance phenomenon from SISR \cite{deville2005two}, has previously been addressed using a machine learning approach based on physics-unaware reservoir computing \cite{hramovForecastingCoherenceResonance2024}, a paradigm fundamentally different from PINNs. In contrast to previous studies that rely purely on data-driven learning or deterministic PINNs, we introduce a Physics-Informed Neural Network based on a Noise-Augmented State Predictor (NASP) that explicitly incorporates stochastic forcing into the state-transition map. This formulation embeds both stochastic differential dynamics and asymptotic escape-time constraints from Kramers’ theory directly into the composite loss. The PINN therefore learns noise-aware slow–fast dynamics from relatively short trajectories while respecting the underlying physics of noise-induced escapes. To our knowledge, this work presents the first PINN framework capable of modeling and predicting SISR, demonstrated here in the stochastic FHN neuron, although the method is in principle applicable to a broad class of multiscale stochastic systems.

Our results demonstrate that the physics-informed framework can accurately capture the
onset and degree of SISR from simulated data while significantly reducing computational
cost compared to traditional numerical simulations. Although the FHN model can be solved
numerically via standard stochastic integrators, our aim is not mere simulation but to develop
a data-driven surrogate capable of reproducing both short-term dynamics and long-term
stochastic statistics from limited trajectory data.  Furthermore, by comparing different
configurations of the loss functional, we show that embedding physical knowledge directly
into the learning process enhances both interpretability and generalization. The proposed
method thus establishes a general paradigm for integrating stochastic dynamical theory with
neural network learning—extending the reach of PINNs to complex noise-driven phenomena
such as self-induced stochastic resonance.

The remainder of this paper is organized as follows. In Sect.~\ref{sec:math_model_dynamics}, we present the mathematical formulation of the stochastic FHN model and identify its excitable regime. Sect.~\ref{SISR_dynamical_system_approach} revisits SISR from a dynamical systems perspective, emphasizing the roles of timescale separation and noise-induced transitions. Sect.~\ref{sec:numerics} quantifies the influence of excitability and timescale parameters on SISR using numerical simulations. Sect.~\ref{sec:sisr_Pinn} introduces the proposed physics-informed neural network framework, detailing its architecture, training strategy, and implementation for the stochastic FHN system. Finally, Sect.~\ref{sec:conclusion} summarizes the findings and discusses future directions for integrating physics-based learning and stochastic dynamics.

\section{Mathematical model and excitable regime}
\label{sec:math_model_dynamics}

\subsection{Mathematical model}
\label{Mathematical_model}

We consider a single FitzHugh--Nagumo (FHN) neuron, a canonical model for studying neuronal excitability and spike generation~\cite{Fitzhugh-1960a,fitzhughImpulsesPhysiologicalStates1961,nagumoActivePulseTransmission1962}. Its dynamics evolve on a fast timescale \( t \) according to the following set of stochastic differential equations (SDEs):
\begin{equation}\label{eq:1}
\begin{split}
\left\{
\begin{array}{lcl}
\frac{dv_{t}}{dt} &=& f(v_{t},w_{t}) + \sigma \eta_t, \\[1.0mm]
\frac{dw_{t}}{dt} &=& g(v_{t},w_{t}),
\end{array}
\right.
\end{split}
\end{equation}
where, for any point \((v,w) \), the deterministic vector field \(\bigl[f(v,w),\,g(v,w)\bigr]^{\top} \) is defined as
\begin{equation}\label{eq:1a}
\begin{split}
\left\{
\begin{array}{lcl}
f(v,w) &=& v(a-v)(v-1) - w, \\[1.0mm]
g(v,w) &=& \varepsilon(bv - cw),
\end{array}
\right.
\end{split}
\end{equation}
where the fast variable \( v_t \in \mathbb{R} \) denotes the membrane potential and the slow variable \( w_t \in \mathbb{R} \) represents the recovery current. The stochastic term \( \eta_t \) is a one-dimensional Brownian motion with zero mean, intensity \( \sigma \ge 0 \), and autocorrelation \( \langle \eta(t), \eta(t') \rangle = \sigma^2 \delta(t - t') \).

The small parameter \( \varepsilon \), defined as \( 0<\varepsilon := \tau/t \ll 1 \), quantifies the separation between the slow (\( \tau \)) and fast (\( t \)) timescales, ensuring that \( v \) evolves much faster than \( w \). Biophysically, \( \varepsilon \) reflects sodium channel kinetics and strongly influences the shape of the action potential~\cite{xu2014parameters}. The excitability threshold, controlled by the parameter \( a \), modulates the fast subsystem’s dynamics due to sodium current,  and typically lies within \( 0 < a < 1 \)~\cite{xu2014parameters}. The parameters \( b > 0 \) and \( c > 0 \) are constants.

\subsection{Deterministic dynamics in the excitable regime}
\label{Deterministic}

In the deterministic limit ($\sigma = 0$), the FHN neuron in Eq.~\eqref{eq:1} possesses a unique and stable fixed point, corresponding to a quiescent, non-oscillatory state. This defines the \textit{excitable regime}, where trajectories starting within the basin of attraction of the fixed point may exhibit at most one large excursion (spike) before returning to equilibrium~\cite{izhikevich2000neural}. Unlike the oscillatory regime, where a stable limit cycle supports continuous spiking, the excitable regime forms the deterministic foundation for the emergence of SISR.

During SISR, coherent spiking arises exclusively from stochastic perturbations and does not involve a bifurcation to a limit cycle. Thus, the system parameters may be chosen far from the Hopf bifurcation point. In this setting, it is the \textit{interaction between intrinsic timescales and stochastic forcing}—rather than proximity to instability—that generates sustained rhythmic activity. This fundamental distinction separates SISR from coherence resonance, which requires parameters near a bifurcation threshold so that weak noise merely triggers an impending deterministic transition~\cite{yamakouControlCoherenceResonance2019,deville2005two}.

We now determine the excitable regime of the FHN model in Eq.~\eqref{eq:1}, defined by the existence of a unique, stable equilibrium point. These fixed points $(v_e,w_e) \in E$ represent the neuron’s rest states, determined by the intersection of the nullclines:
\begin{equation}\label{eq:2}
E = \big\{ (v_e,w_e) \in \mathbb{R}^2 : v_e(a-v_e)(v_e-1)-w_e=0,\quad \varepsilon(bv_e-cw_e)=0 \big\}.
\end{equation}
The model admits at most three fixed points, depending on the sign of the discriminant
\begin{equation}\label{eq:3}
\Delta = (a + 1)^2 - 4 \left(a + \frac{b}{c}\right).
\end{equation}
We focus on the case $\Delta < 0$, for which the system has a unique equilibrium at $(v_e, w_e) = (0, 0)$. By selecting parameters satisfying this condition, the deterministic FHN neuron operates in the excitable regime—providing the baseline dynamics for SISR.

The fixed point $(v_e, w_e) = (0, 0)$ is unique and stable if and only if the following conditions hold:
\begin{equation}\label{eq:4}
\begin{split}
\left\{
\begin{array}{lcl}
\Delta < 0 \Longleftrightarrow \frac{(a - 1)^2}{4} < \frac{b}{c}, \\[1.0mm]
\text{tr}(J) < 0 \Longleftrightarrow a + \varepsilon c > 0, \\[1.0mm]
\text{det}(J) > 0 \Longleftrightarrow \varepsilon(ac + b) > 0,
\end{array}
\right.
\end{split}
\end{equation}
where $J$ is the Jacobian matrix evaluated at $(v_e, w_e) = (0, 0)$.

Throughout this paper, we fix $b = 1.0$ and $c = 2.0$. Under these parameters, the system lies in the excitable regime if the excitability parameter $a$ satisfies
\begin{equation}\label{eq:5}
    0 < a < 1 + \sqrt{2}.
\end{equation}
Choosing $a$ above this upper bound violates the uniqueness condition of the fixed point, whereas $a$ below the lower bound induces self-sustained oscillations via a Hopf bifurcation. We therefore restrict attention to the interval $a \in (0, 1 + \sqrt{2})$.

\begin{figure}
\centering
\includegraphics[width=7.5cm,height=5.0cm]{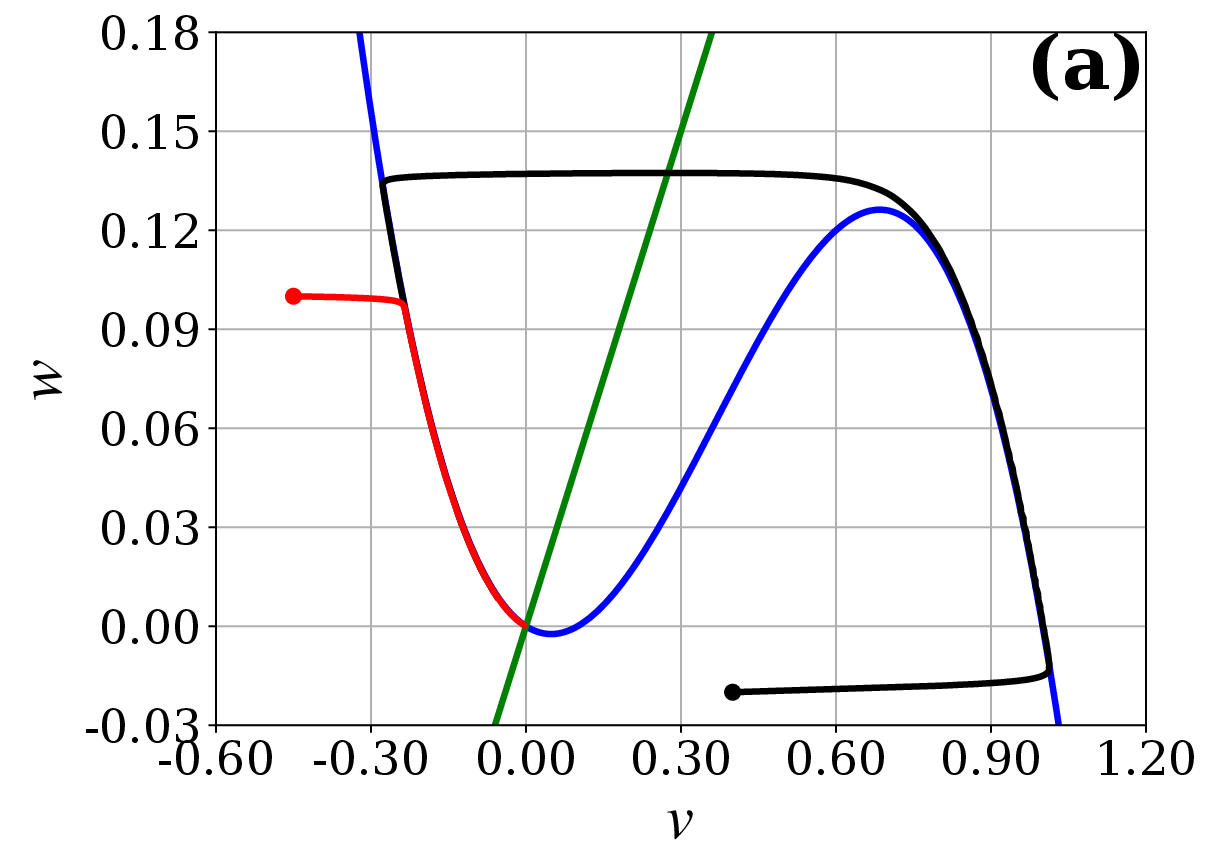}\includegraphics[width=7.5cm,height=5.0cm]{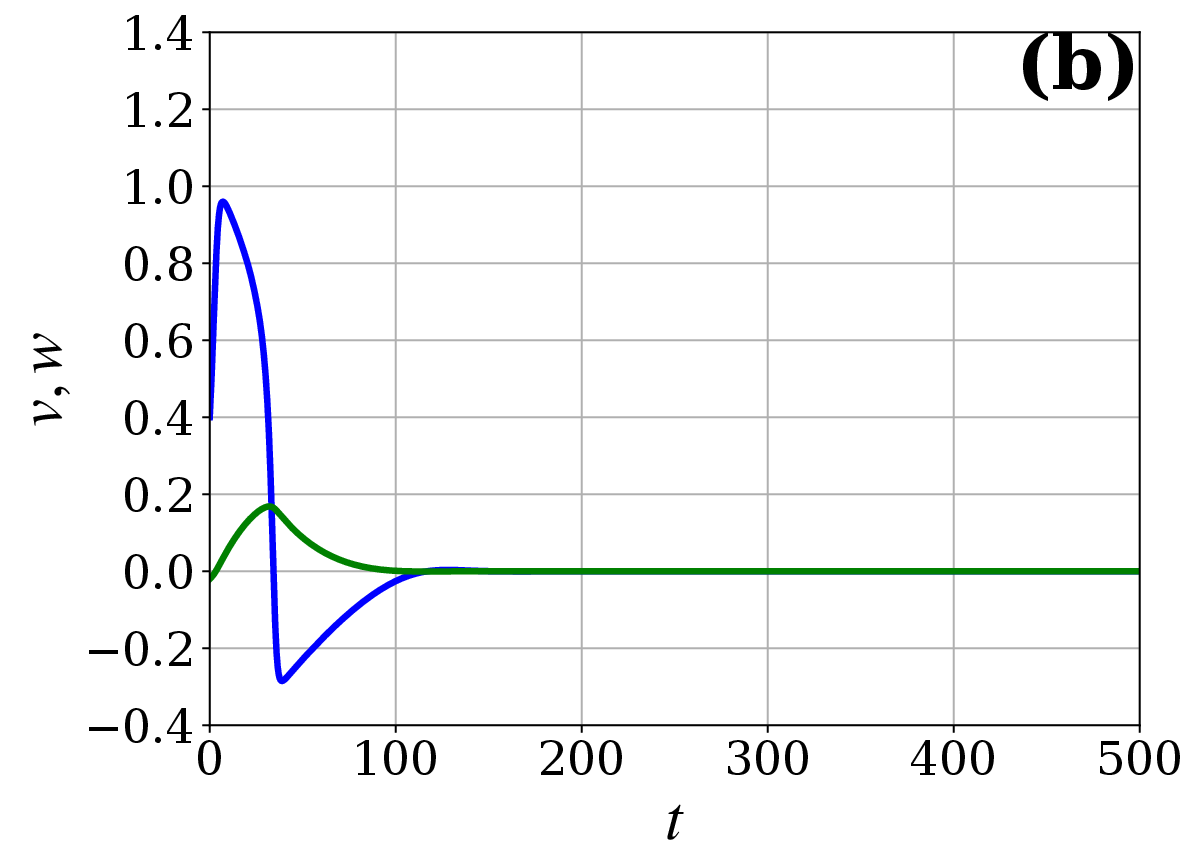}
\caption{Excitable regime of the FHN model. 
(a) Phase portrait with two trajectories from different initial conditions converging to the fixed point $(v,w)=(0,0)$. 
(b) Time series of $v(t)$ (blue) and $w(t)$ (green) for the black trajectory in (a). 
Parameters: $a=0.1$, $b=1.0$, $c=2.0$.}
    \label{fig:1}
\end{figure}

Figure~\ref{fig:1}(a) illustrates the phase portrait of the FHN system in this excitable regime, where trajectories starting from different initial conditions converge to the stable fixed point $(v_e, w_e) = (0, 0)$. The corresponding time series of $v(t)$ and $w(t)$ in Fig.~\ref{fig:1}(b) confirms the absence of sustained oscillations in the deterministic setting.

\section{SISR: The physics approach}
\label{SISR_dynamical_system_approach}

In this section, we analyze SISR from the perspective of nonlinear dynamics, viewing it as a noise-driven phenomenon emerging from the interplay between fast--slow dynamics and stochastic perturbations in excitable systems. We characterize the potential landscape, energy barriers, and noise-induced transitions that underpin the coherent spiking behavior of SISR. This physics-based understanding provides the theoretical foundation for the physics-informed neural network framework developed in the next section.

We consider the stochastic FHN neuron of Eq.~\eqref{eq:1} in the excitable regime, $0<a<1+\sqrt{2}$, where the deterministic dynamics admit only a stable fixed point. Our goal is to examine how stochastic forcing of amplitude $\sigma$ generates coherent spikes in the absence of any deterministic oscillations.

\begin{figure}
\includegraphics[width=5.0cm,height=4.0cm]{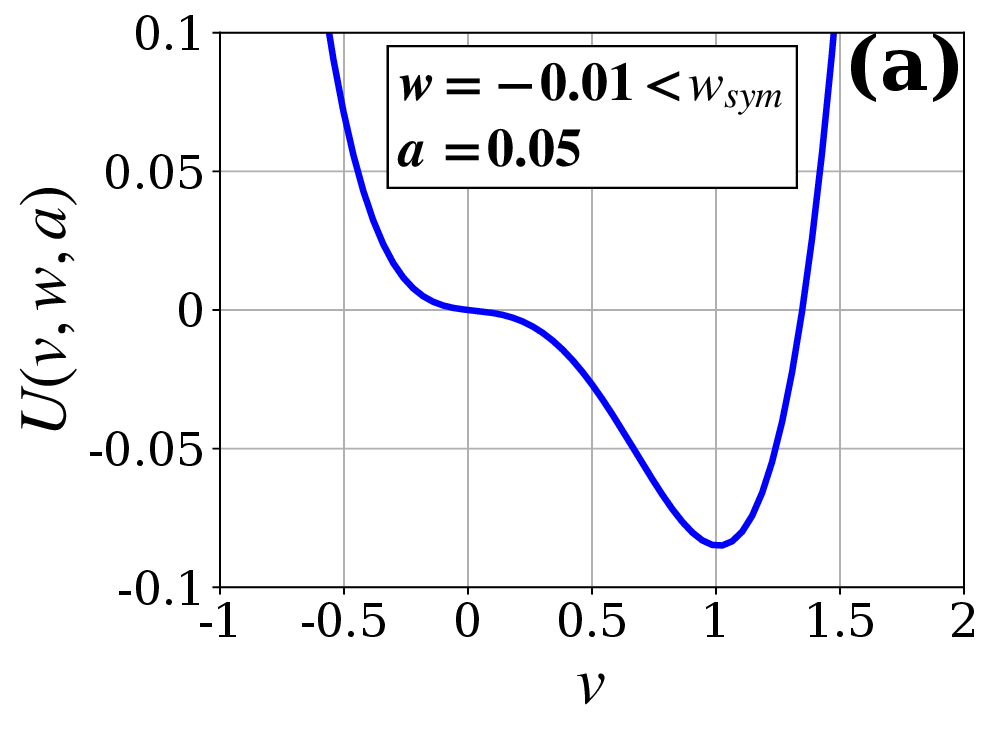}
\includegraphics[width=5.0cm,height=4.0cm]{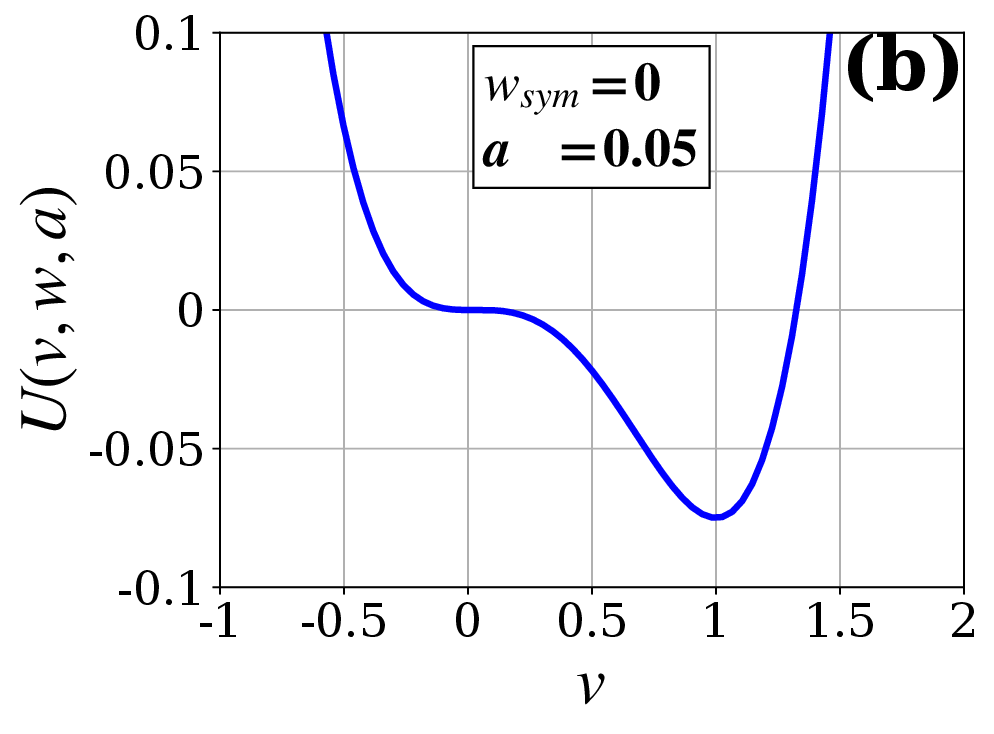}
\includegraphics[width=5.0cm,height=4.0cm]{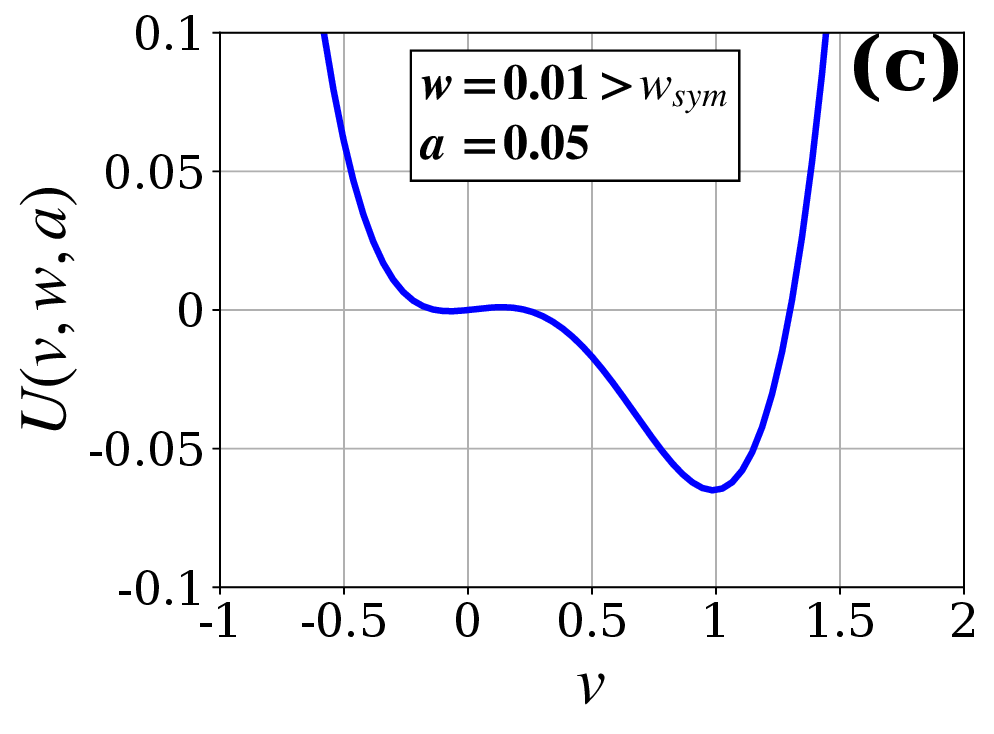}
\includegraphics[width=5.0cm,height=4.0cm]{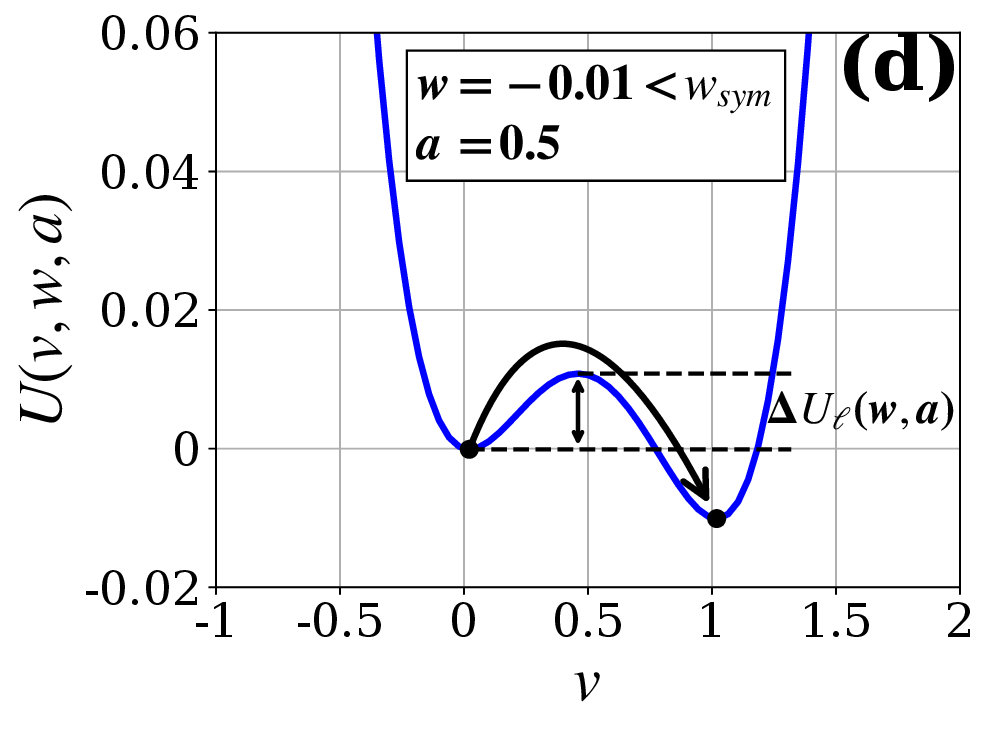}
\includegraphics[width=5.0cm,height=4.0cm]{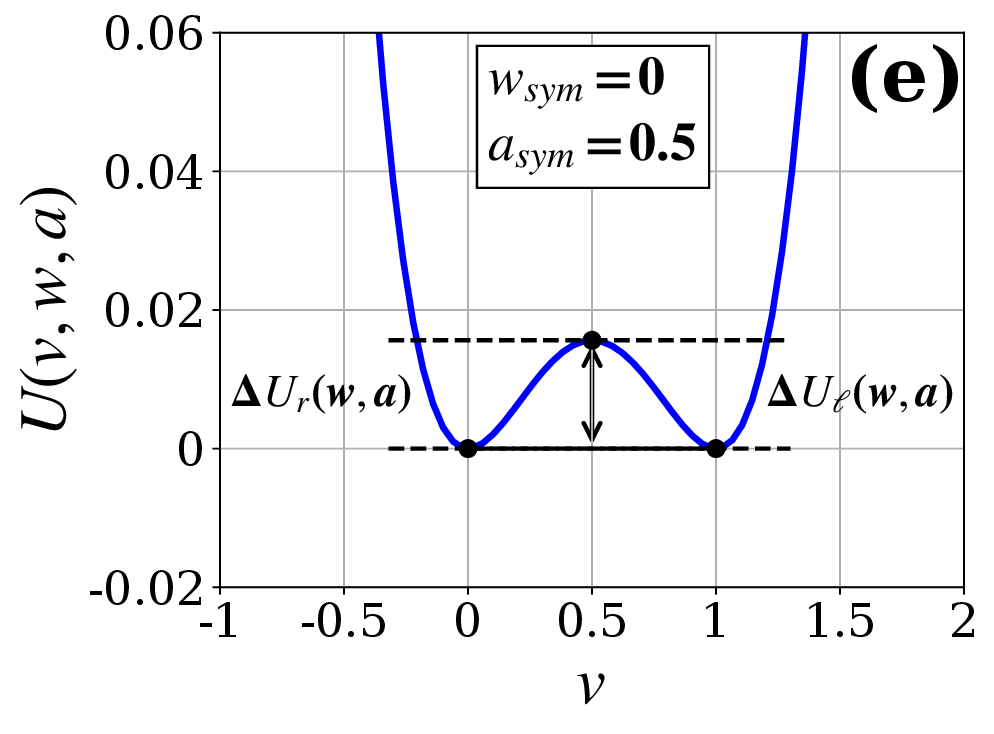}
\includegraphics[width=5.0cm,height=4.0cm]{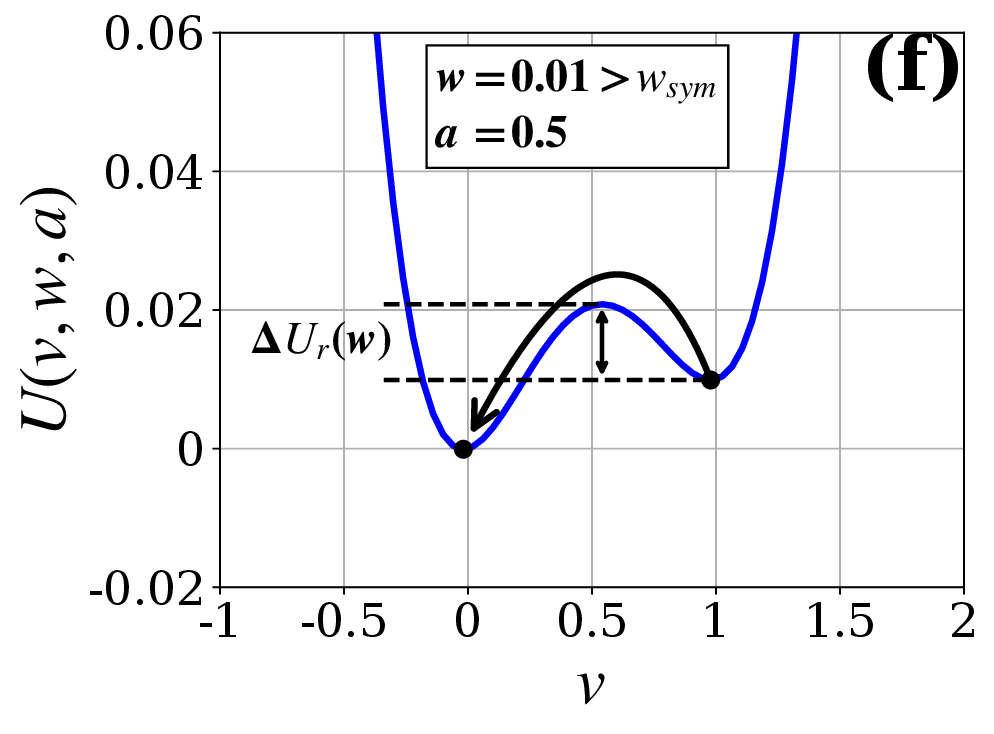}
\includegraphics[width=5.0cm,height=4.0cm]{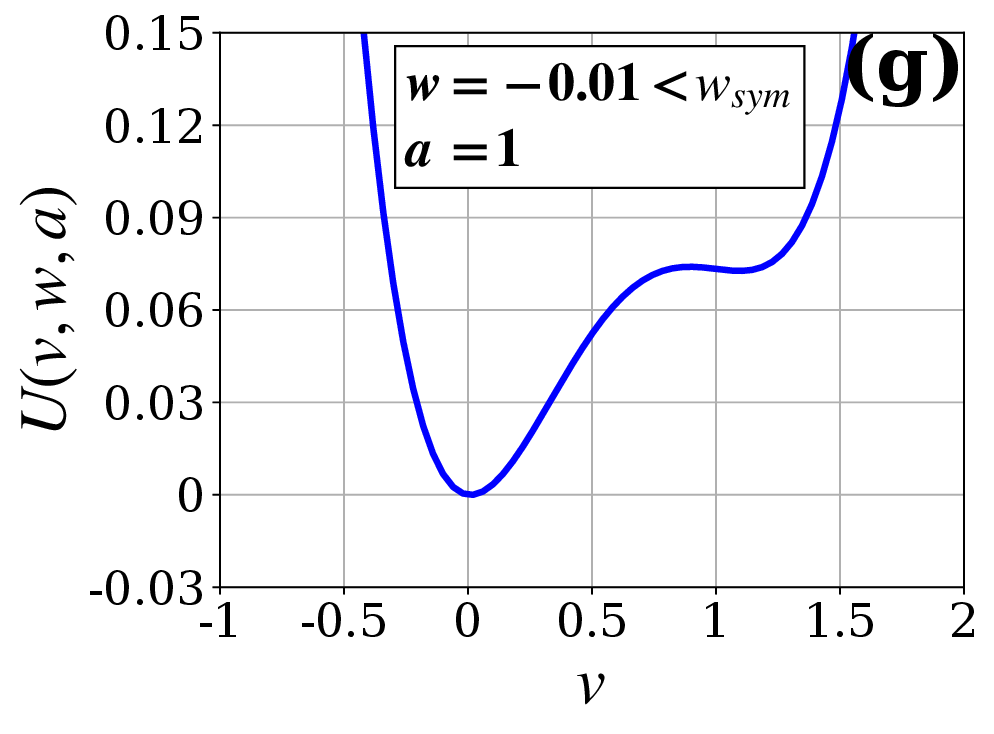}
\includegraphics[width=5.0cm,height=4.0cm]{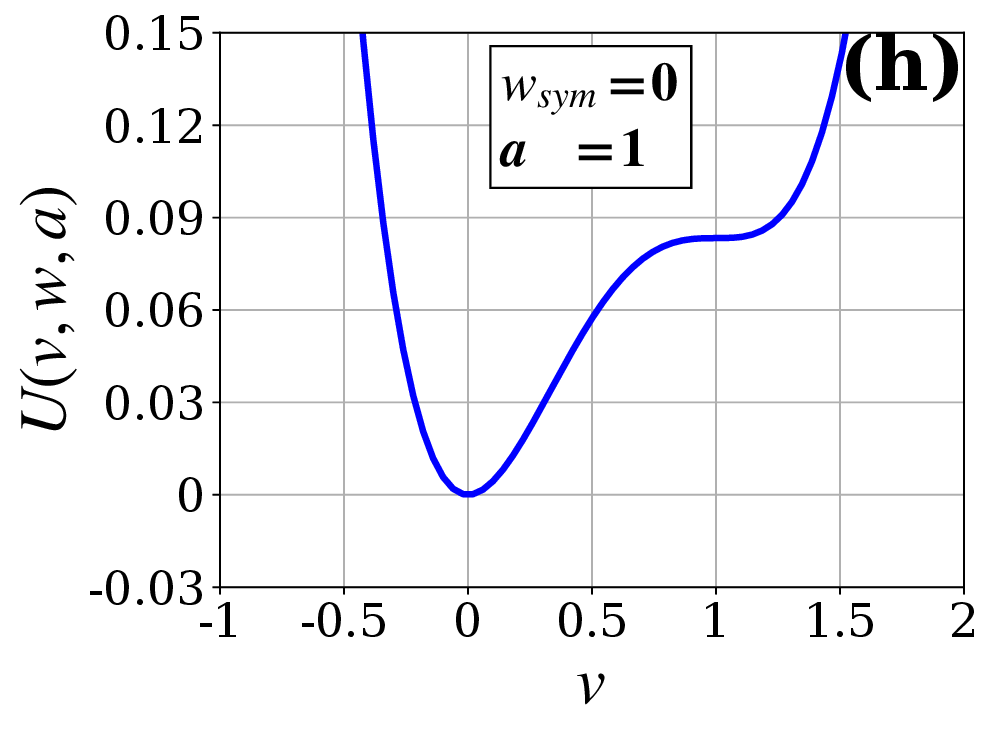}
\includegraphics[width=5.0cm,height=4.0cm]{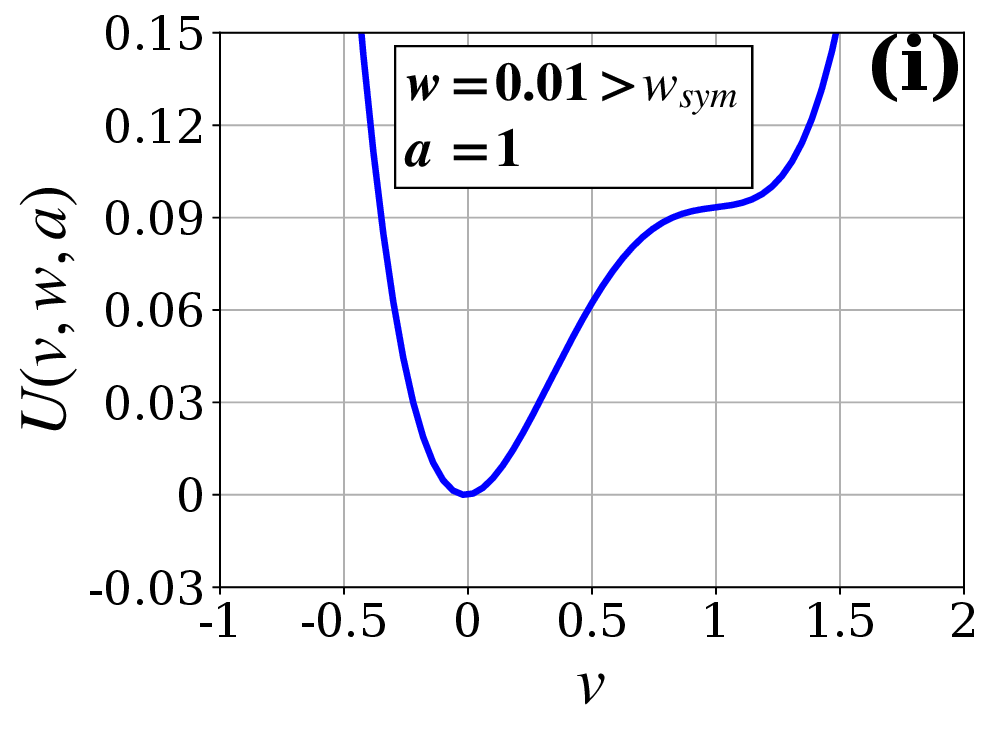}
\caption{Potential landscape $U(v,w,a)$ for different $(a,w)$. The asymmetry of the double-well potential varies with $a$ and $w$: (a)--(c) right asymmetry for small $a$, (e) symmetry at intermediate $a$ for $w=0$, (d),(f) asymmetry at intermediate $a$ for $w\neq0$, and (g)--(i) left asymmetry for large $a$. $\Delta U_{\ell}(w,a)$ and $\Delta U_{r}(w,a)$ denote the left and right potential barriers.}
\label{fig:2}
\end{figure}

In the adiabatic limit $\varepsilon \to 0$, the timescale separation between $v_t$ and $w_t$ becomes large. On the $O(1)$ fast timescale, $w_t$ may be treated as quasi-static, and Eq.~\eqref{eq:1} reduces to the Langevin equation
\begin{equation}\label{eq:6}
\begin{split}
\left\{
\begin{array}{lcl}
dv &=& -\frac{\partial U(v,w,a)}{\partial v}\,dt + \sigma\,dB, \\[1.0mm]
dw &=& 0,
\end{array}
\right.
\end{split}
\end{equation}
with the effective potential
\begin{equation}\label{pot}
U(v,w,a) = \tfrac{1}{4}v^{4} - \tfrac{1}{3}(a+1)v^{3} + \tfrac{1}{2}a v^{2} + v w,
\end{equation}
representing a double-well landscape parametrized by $(a,w)$. The potential shape and asymmetry depend on both parameters, as shown in Fig.~\ref{fig:2}, with $\Delta U_{\ell}(w,a)$ and $\Delta U_{r}(w,a)$ denoting the left and right barrier heights, respectively.

\begin{figure}
\centering
\includegraphics[width=7.0cm,height=5.0cm]{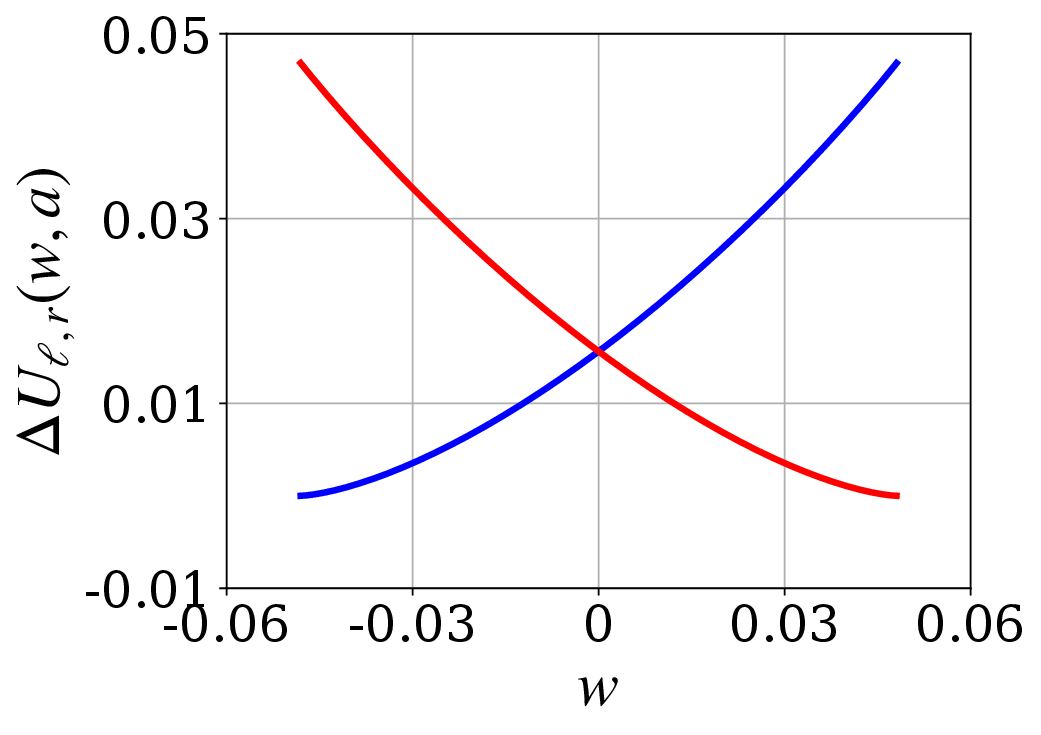}
\caption{Monotonic dependence of the potential barriers $\Delta U_{\ell}(w,a)$ (blue) and 
$\Delta U_{r}(w,a)$ (red) on $w \in [w_{\min}, w_{\max}]=[-0.04775, 0.04775]$ at $a=0.5$.}
\label{fig:3}
\end{figure}

The left ($\Delta U_{\ell}$) and right ($\Delta U_{r}$) energy barriers are defined as
\begin{equation}\label{eq:8}
\begin{split}
\left\{
\begin{array}{lcl}
\Delta U_{\ell}(w,a) &=& U(v_{s}^*(w,a), w) - U(v_{\ell}^*(w,a), w), \\[1.0mm]
\Delta U_{r}(w,a) &=& U(v_{s}^*(w,a), w) - U(v_{r}^*(w,a), w),
\end{array}
\right.
\end{split}
\end{equation}
which vary monotonically with $w$ over $[w_{\min}, w_{\max}]$ (Fig.~\ref{fig:3}).

The extremal points $v_{\ell}^*(w,a)$, $v_{s}^*(w,a)$, and $v_{r}^*(w,a)$ in Eq.~\eqref{eq:8} are the real roots of the cubic nullcline $v[(a-v)(v-1)]-w=0$:
\begin{equation}\label{eq:15}
\begin{split}
\left\{
\begin{array}{lcl}
v_{\ell}^*(w,a) &=&  \frac{a+1}{3} + \frac{2}{3} \sqrt{a^2 - a + 1}\cos\!\Big[\tfrac{1}{3}\arccos\!\Big(\frac{(a+1)^3 - 9 a (a+1)/2 -27 w/2}{(a^2 - a + 1)^{3/2}}\Big) + \tfrac{2\pi}{3}\Big],\\[1.5mm]
v_{s}^*(w,a) &=&  \frac{a+1}{3} + \frac{2}{3} \sqrt{a^2 - a + 1}\cos\!\Big[\tfrac{1}{3}\arccos\!\Big(\frac{(a+1)^3 - 9 a (a+1)/2 -27 w/2}{(a^2 - a + 1)^{3/2}}\Big) - \tfrac{2\pi}{3}\Big],\\[1.5mm]
v_{r}^*(w,a) &=&  \frac{a+1}{3} + \frac{2}{3} \sqrt{a^2 - a + 1}\cos\!\Big[\tfrac{1}{3}\arccos\!\Big(\frac{(a+1)^3 - 9 a (a+1)/2 -27 w/2}{(a^2 - a + 1)^{3/2}}\Big)\Big],
\end{array}
\right.
\end{split}
\end{equation}
ordered as $v_{\ell}^*(w,a) < v_{s}^*(w,a) < v_{r}^*(w,a)$.

Substituting \(v=v_{\ell}^*(w,a)\) and \(v=v_{r}^*(w,a)\) into the $w$-equation of Eq.~\eqref{eq:1} gives
\begin{eqnarray}\label{eq:6c}
\left\{
\begin{array}{lcl}
dw &=& \varepsilon [ bv_{\ell}^*(w,a) - cw ]\,dt,\\[1.0mm]
dw &=& \varepsilon [ bv_{r}^*(w,a) - cw ]\,dt,
\end{array}
\right.
\end{eqnarray}
which describe the slow evolution of $w$ along the left and right stable branches of the $v$-nullcline as $\varepsilon \to 0$.

To analyze stochastic transitions between the potential wells $\mathcal{B}(v_{\ell}^*)$ and $\mathcal{B}(v_{r}^*)$, we adopt the classical analogy of Brownian motion in a double-well potential~\cite{kramersBrownianMotionField1940,risken1991fokker}. In the weak-noise limit ($0<\sigma\ll1$), rare escapes are governed by Kramers’ law~\cite{berglund2006noise,kramersBrownianMotionField1940}, where the probability density $p(v,t)$ satisfies the Fokker--Planck equation
\begin{equation}\label{eq:11}
\frac{\partial p(v,t)}{\partial t} = -\frac{\partial}{\partial v}\!\Big[U(v,w)p(v,t)\Big] 
+ \frac{\sigma^2}{2}\frac{\partial^2 p(v,t)}{\partial v^2},
\end{equation}
subject to
\begin{equation}\label{eq:12}
\begin{split}
\left\{
\begin{array}{lcl}
p(v,t_0|v_0,t_0)=\delta(v-v_0),\\[1.0mm]
\lim\limits_{v\to\pm\infty}p(v,t)=\lim\limits_{v\to\pm\infty}\partial_v p(v,t)=0.
\end{array}\right.
\end{split}
\end{equation}

The stationary probability flux $j_0$ in the limit $\sigma \to 0$ yields the escape rate in Arrhenius form~\cite{berglund2006noise,risken1991fokker}:
\begin{equation}\label{eq:25}
k_{\ell,r}=c_{\ell,r}\exp\!\Big(-\frac{2\Delta U_{\ell,r}(w,a)}{\sigma^2}\Big),
\end{equation}
where the prefactors
\begin{equation}\label{eq:27}
c_{\ell,r}=\frac{1}{2\pi}\Big(U''(v_{\ell,r}^*)\,|U''(v_{s}^*)|\Big)^{1/2},
\end{equation}
are sub-exponential in the weak noise limit ($0<\sigma\ll1$) so that the mean escape times (stochastic timescales) given by \cite{kramersBrownianMotionField1940}
\begin{equation}
\tau_{\ell,r} = \frac{1}{k_{\ell,r}} = c_{\ell,r}\exp\!\Big(\frac{2\Delta U_{\ell,r}(w,a)}{\sigma^2}\Big),
\end{equation}
become exponentially long as $\sigma \to 0$.

From Eq.~\eqref{eq:1}, the deterministic timescale of motion along the stable nullcline branches is $\varepsilon^{-1}$~\cite{muratovSelfinducedStochasticResonance2005,yamakouCoherentNeuralOscillations2018}. For $\sigma=0$, trajectories follow these branches adiabatically and converge to $(v_e,w_e)=(0,0)$. When $\sigma\neq0$, noise induces escapes between branches, corresponding to spike generation. SISR arises precisely when the deterministic timescale $\varepsilon^{-1}$ and the stochastic timescale $\tau_{\ell,r}$ match at the unique exit points $w_{\ell}$ and $w_r$~\cite{muratovSelfinducedStochasticResonance2005,yamakouCoherentNeuralOscillations2018,deville2007nontrivial}.

If $\varepsilon^{-1}<\tau_{\ell,r}$, trajectories remain near $(v_e,w_e)$, producing rare, incoherent spikes; if $\varepsilon^{-1}>\tau_{\ell,r}$, escapes occur too frequently and irregularly. Coherent oscillations therefore arise only when these two timescales synchronize, \textit{i.e.,} when $\varepsilon^{-1}=\tau_{\ell,r}$ at the exit locations $w=w_{\ell,r}$. 

This heuristic criterion can be made rigorous using large-deviation theory~\cite{freidlin2001stochastic,freidlin2001stable} and Kramers’ escape-rate law~\cite{kramersBrownianMotionField1940}, which translates the timescale matching into explicit conditions on barrier heights and parameters. For slow–fast systems of the form Eq.~\eqref{eq:1}, the resulting SISR conditions are~\cite{muratovSelfinducedStochasticResonance2005,yamakouCoherentNeuralOscillations2018,deville2007nontrivial}:
\begin{eqnarray}\label{eq:15}
\begin{split}
\left\{
\begin{array}{lcl}
v_e<v_{\min} 	\Longleftrightarrow a>0,\\[1.0mm]
w_e<w_{\ell}	\Longleftrightarrow\lim\limits_{(\varepsilon,\sigma)\to (0,0)}\Big[\tfrac{1}{2}\sigma^2\log(\varepsilon^{-1})\Big] =O(1),\\[1.5mm]
\lim\limits_{(\varepsilon,\sigma)\to (0,0)}\Big[\tfrac{1}{2}\sigma^2\log(\varepsilon^{-1})\Big] =
\begin{cases}
\Delta U_{\ell}(w_{\ell},a), \quad w<0,\\
\Delta U_{r}(w_r,a), \quad w>0,
\end{cases}\\[2.5mm]
\Delta U_{\ell}(w,a) \nearrow w\in[w_{\min},w_{\max}], \quad 
\Delta U_{r}(w,a) \searrow w\in[w_{\min},w_{\max}],
\end{array}\right.
\end{split}
\end{eqnarray}
where $(v_{\min},w_{\min})$ and $(v_{\max},w_{\max})$ are the extrema of the $v$-nullcline, $(v_e,w_e)=(0,0)$ is the unique fixed point, and $w_{\ell}, w_r$ are the escape points on the left and right branches of the $v$-nulllcine satisfying Eq.~\eqref{eq:6c}. 

In Eq. \eqref{eq:15}, the first condition guarantees uniqueness and stability of the fixed point; the second ensures that escape from the left stable branch of the $v$-nullcline occurs at $w=w_{\ell}$ before trajectories are trapped in the basin of attraction of the fixed point $(v_e,w_e)$; the third ensures almost-sure escape from the left (right) branch of the $v$-nullcline at $w=w_{\ell}$ ($w_r$); and the fourth ensures uniqueness of $w_{\ell,r}$ through the monotonicity of $\Delta U_{\ell,r}(w,a)$ (Fig. \ref{fig:3}), leading to periodic, coherent spiking. The coherence of the spike train depends on how accurately the third condition is fulfilled, distinguishing SISR from coherence resonance, which instead requires parameter tuning near a bifurcation~\cite{deville2005two,hramovForecastingCoherenceResonance2024}.

\section{Numerical results: effect of excitability and timescale separation on SISR}
\label{sec:numerics}
To quantify the degree of SISR as a function of the excitability parameter $a$ and the timescale separation $\varepsilon$, we use the \textit{coefficient of variation} ($\mathrm{CV}$) of interspike intervals (ISIs)~\cite{pikovskyCoherenceResonanceNoisedriven1997}. Biophysically, $\mathrm{CV}$ reflects the temporal precision of spike generation~\cite{pei1996noise} and is therefore more informative than global spectral measures. For consecutive spike times $t^k$ and $t^{k+1}$, with ISI$^k = t^{k+1}-t^k$, we define
\begin{equation}\label{eq:cv}
\mathrm{CV}=\dfrac{\sqrt{\langle \text{ISI}^2\rangle-\langle \text{ISI}\rangle^2}}{\langle \text{ISI}\rangle},
\end{equation}
where $\langle \text{ISI}\rangle$ and $\langle \text{ISI}^2\rangle$ are the mean and mean-squared ISIs, respectively. Spikes are detected when the membrane potential $v(t)$ crosses the threshold $v_{\mathrm{th}}=0.4$ from below.

A value $\mathrm{CV}=1$ indicates a Poisson-like (incoherent) spike train, \marius{whereas} $\mathrm{CV}>1$ corresponds to super-Poissonian irregularity~\cite{kurrer1995noise}. In both cases, the system is far from the synchronization condition of Eq.~\eqref{eq:15}. Conversely, as $\mathrm{CV}\to0$, the ISIs become highly regular and the SISR condition is increasingly well satisfied. Perfectly periodic spiking corresponds to $\mathrm{CV}=0$, \marius{and} small but nonzero values of $\mathrm{CV}$ quantify strong SISR. Furthermore, we note that $\mathrm{CV}$ becomes mathematically meaningful 
(\textit{i.e.,} informative about spike-train coherence) only when at least three ISIs 
are available; however, a reliable estimate typically requires many more ISIs 
(on the order of $\sim 10$ for a coarse estimate).

\begin{figure}[h]
\centering
\includegraphics[width=5.0cm,height=4.0cm]{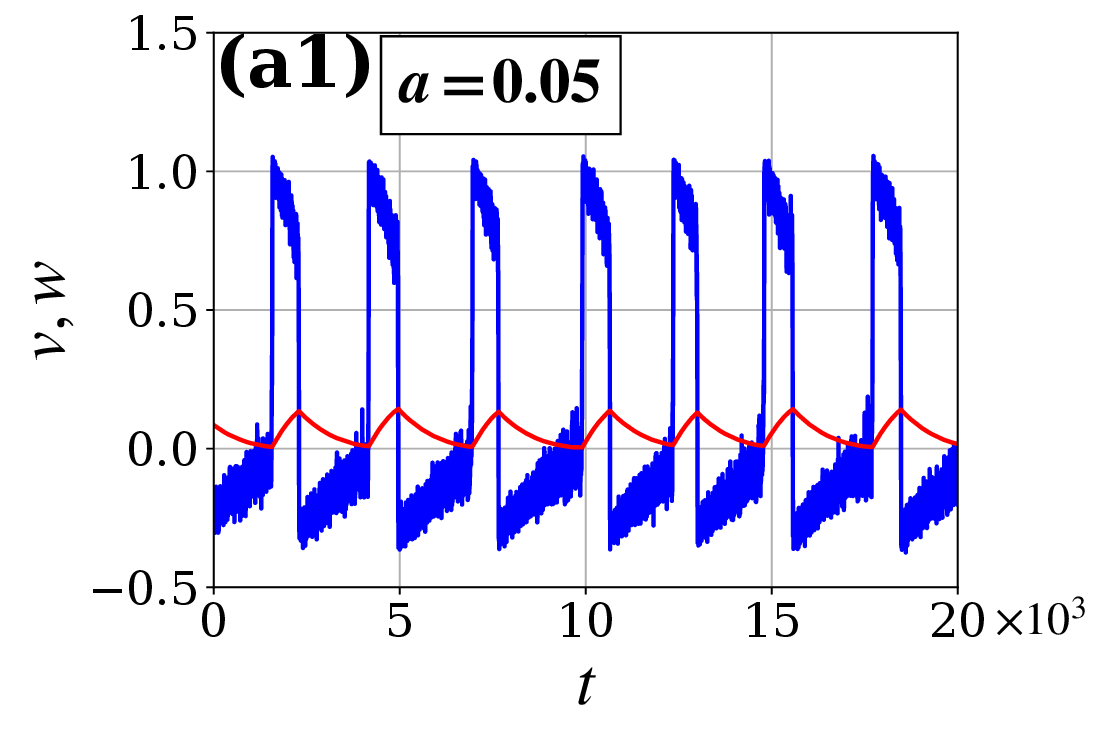}
\includegraphics[width=5.0cm,height=4.0cm]{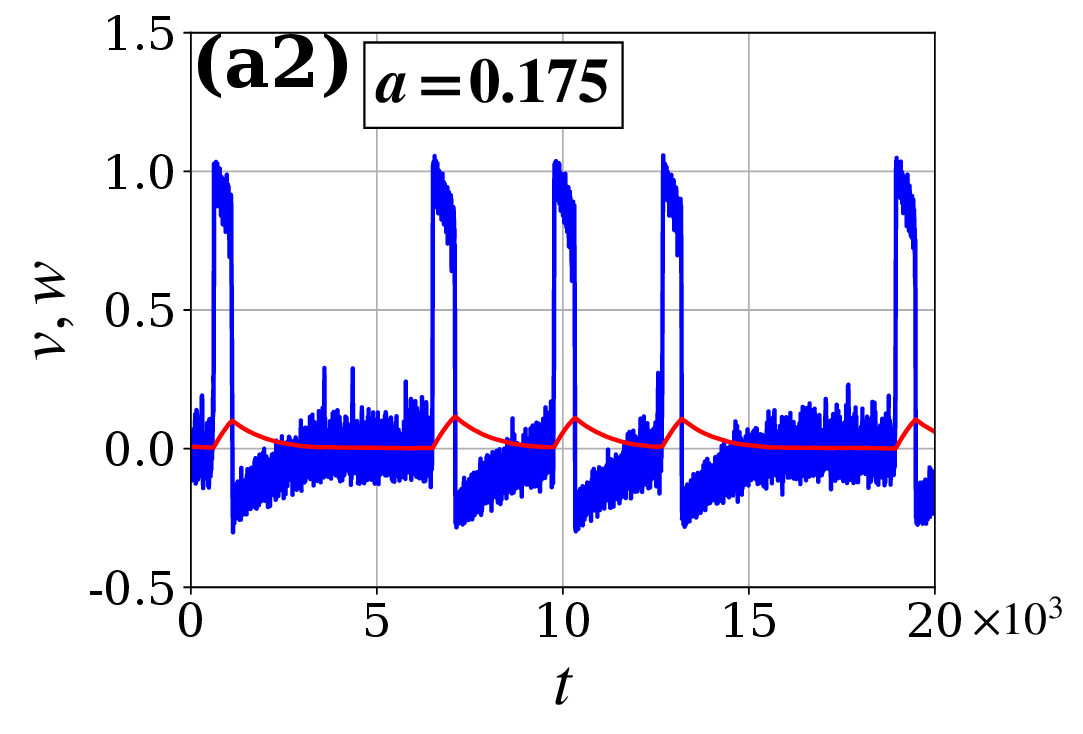}
\includegraphics[width=5.0cm,height=4.0cm]{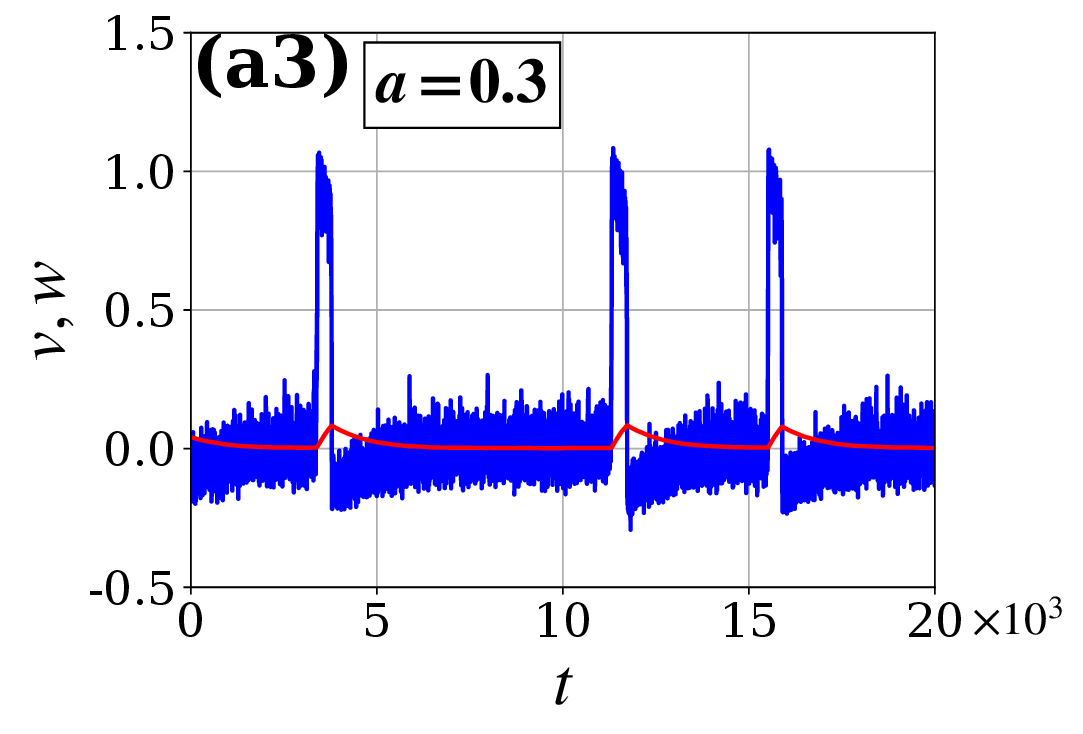}\\
\includegraphics[width=5.0cm,height=4.0cm]{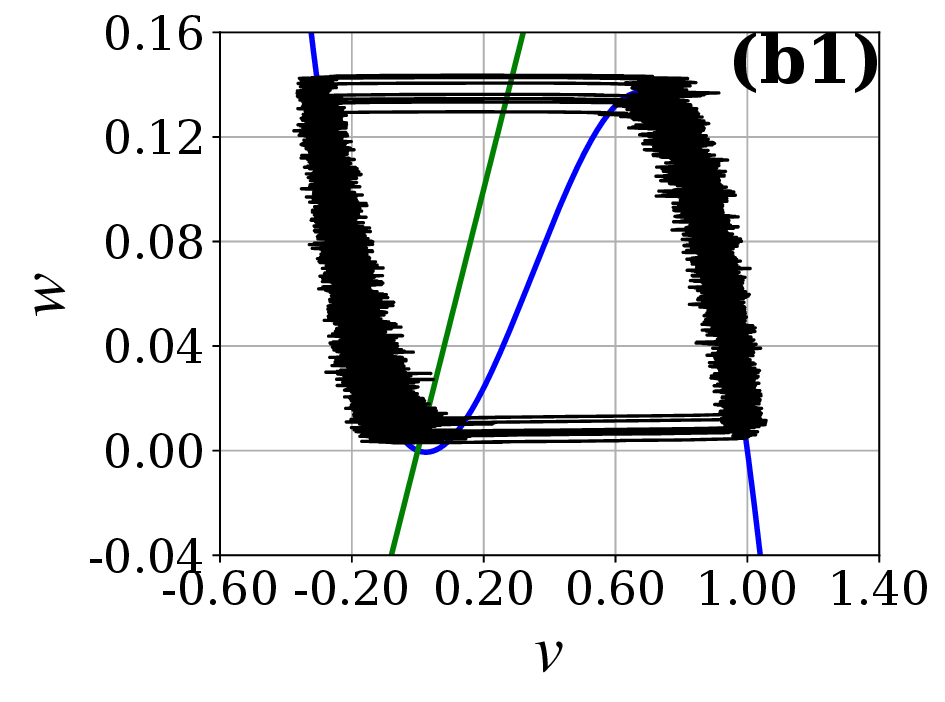}
\includegraphics[width=5.0cm,height=4.0cm]{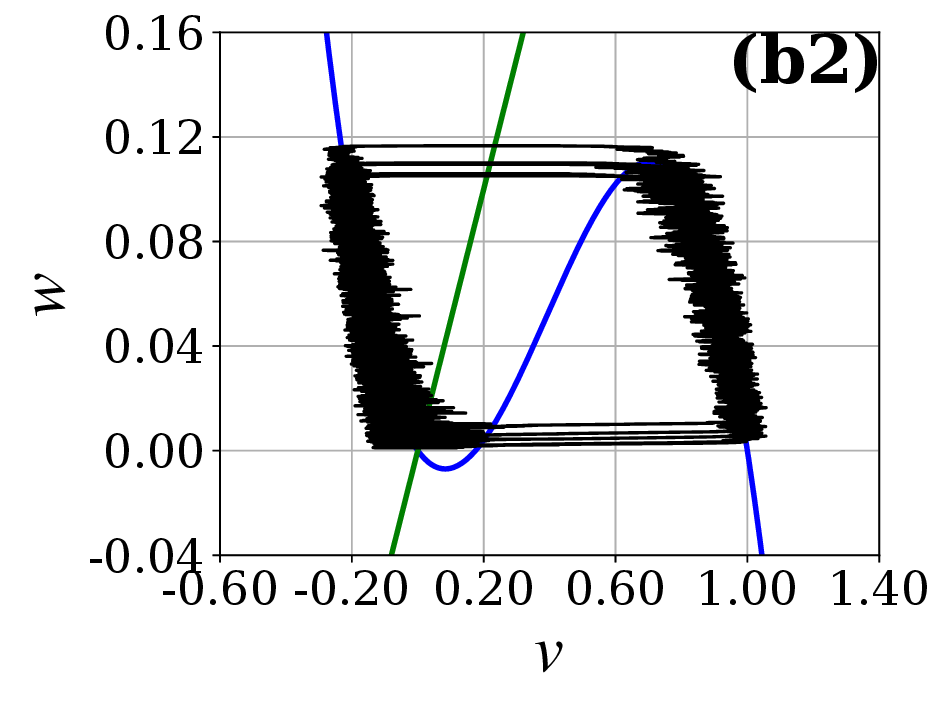}
\includegraphics[width=5.0cm,height=4.0cm]{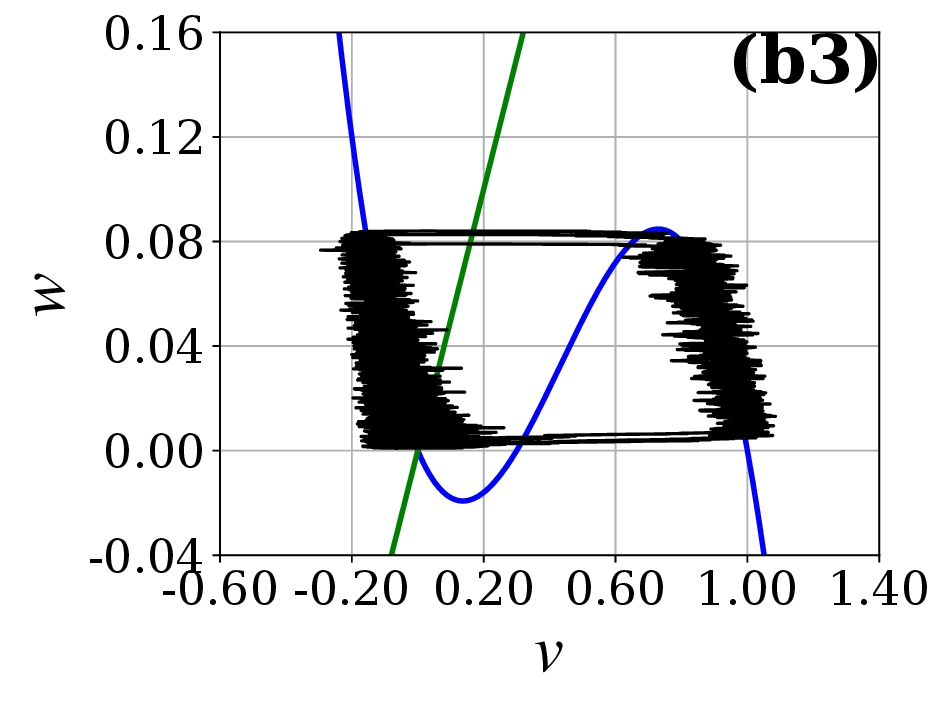}\\
\includegraphics[width=6.0cm,height=4.0cm]{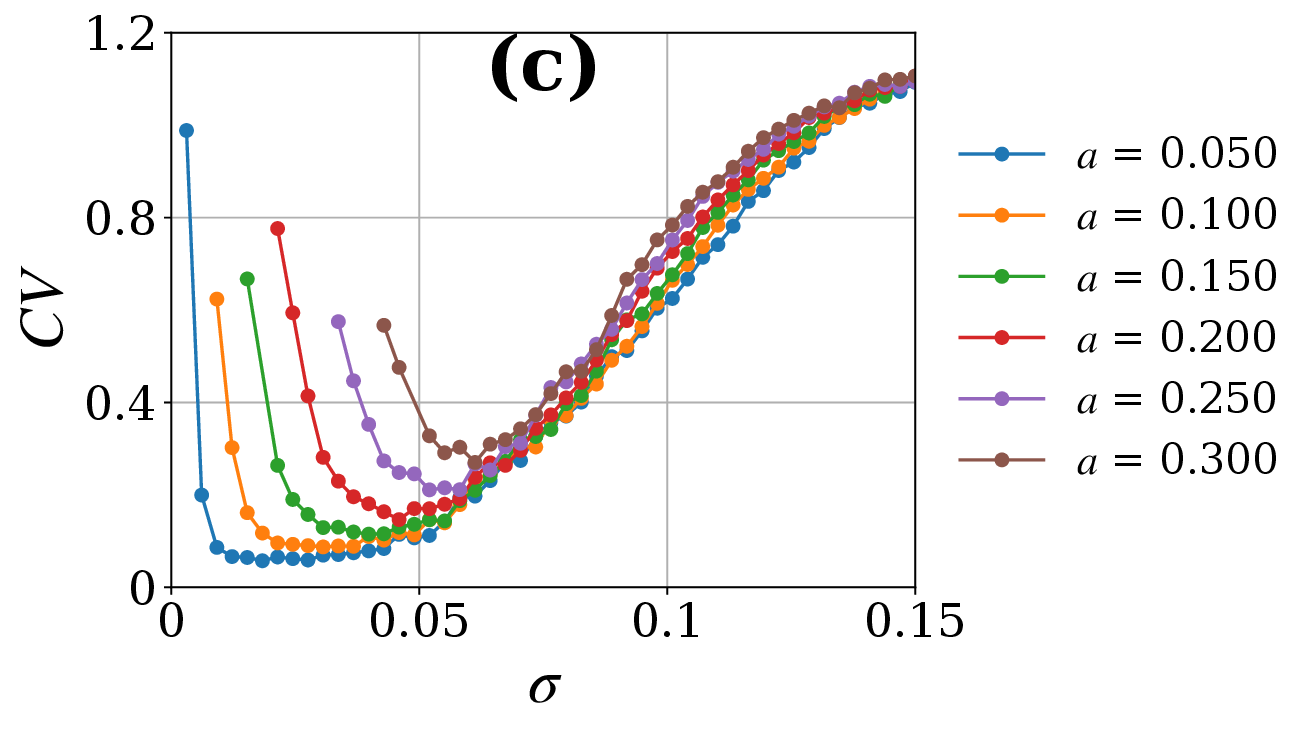}
\caption{Effect of the excitability parameter $a$ on SISR. (a1)--(a3) Time series of
$v(t)$ (blue) and $w(t)$ (red) at $\sigma=0.05$. (b1)--(b3) Corresponding phase portraits.
(c) $\mathrm{CV}$ vs. noise intensity $\sigma$ for different $a$. Smaller $a$ values yield stronger SISR at weaker noise intensities.
Parameters: $\varepsilon=0.00025$, $b=1.0$, $c=2.0$.}
\label{fig:4}
\end{figure}

\begin{figure}
\centering
\includegraphics[width=5.0cm,height=4.0cm]{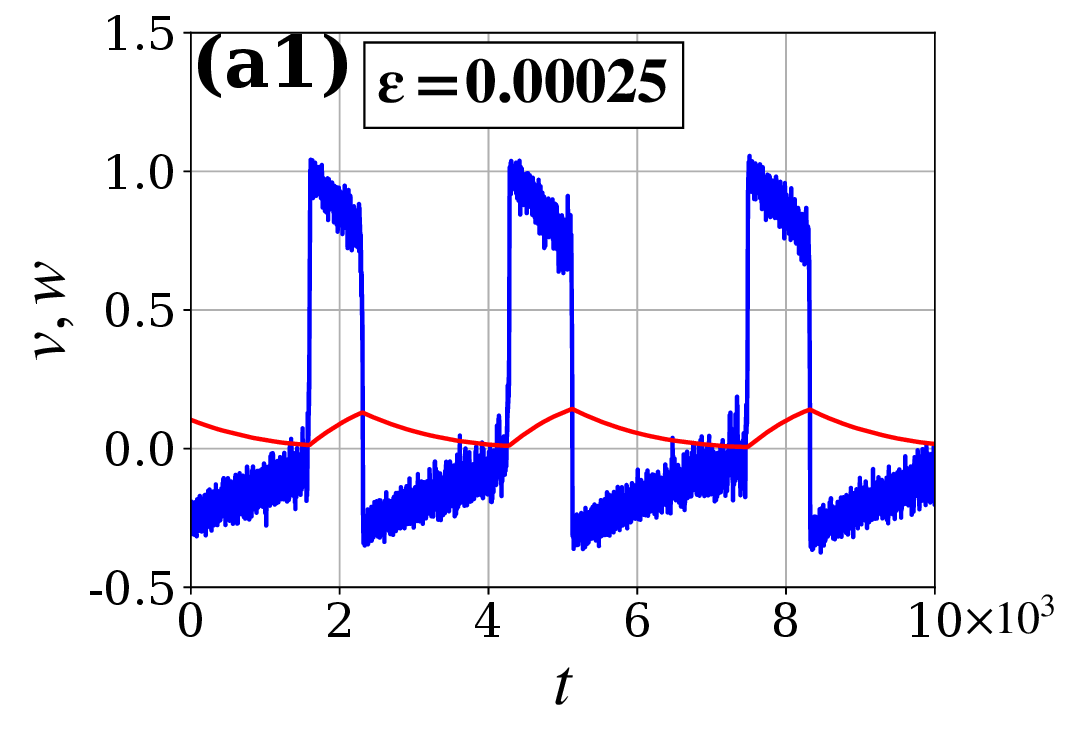}
\includegraphics[width=5.0cm,height=4.0cm]{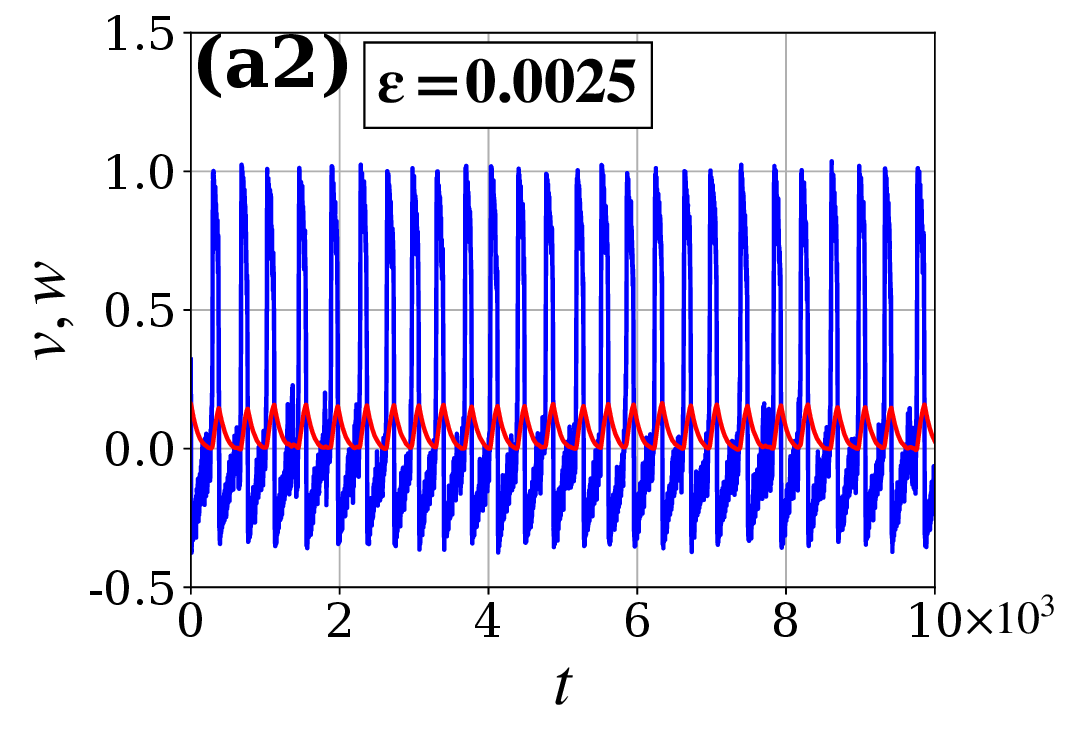}
\includegraphics[width=5.0cm,height=4.0cm]{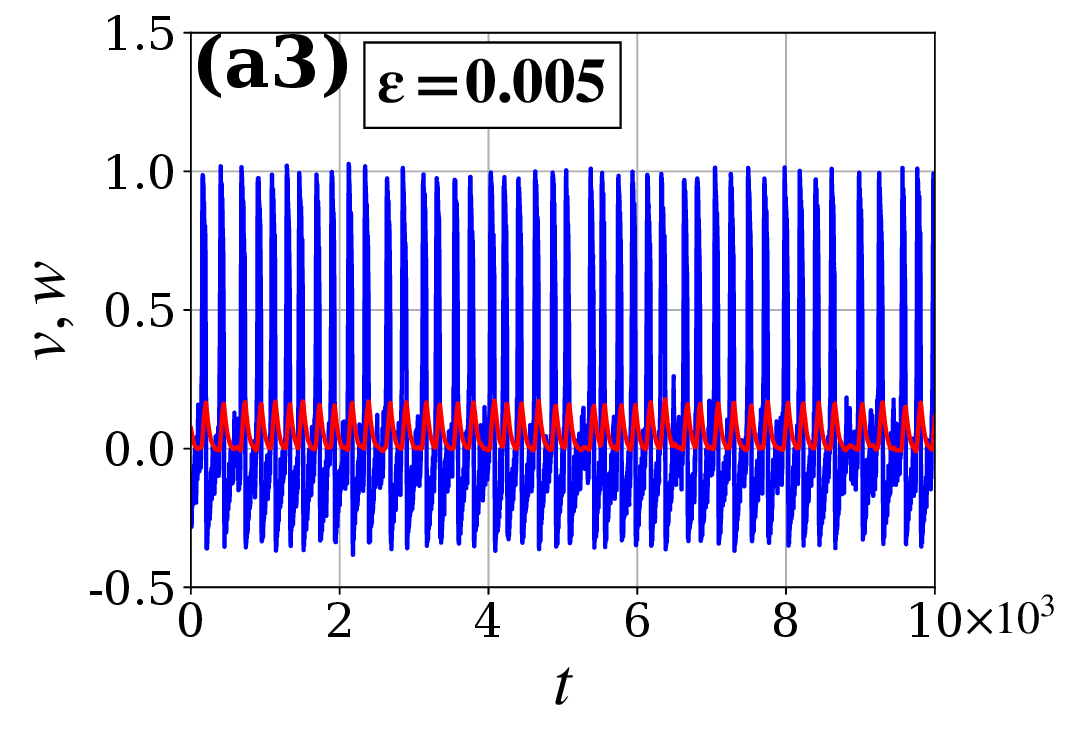}\\
\includegraphics[width=5.0cm,height=4.0cm]{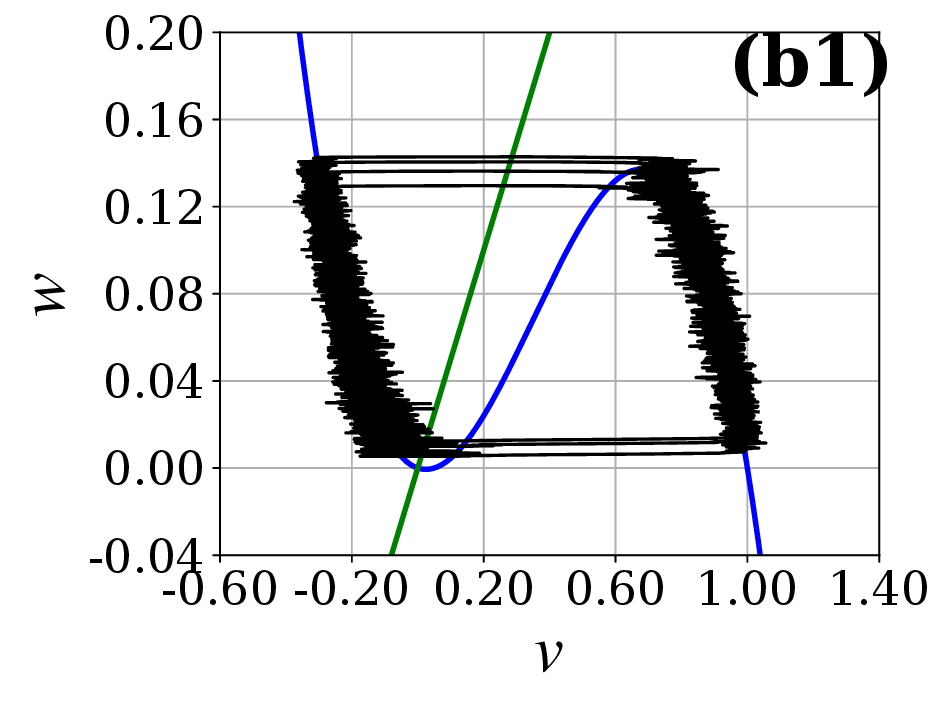}
\includegraphics[width=5.0cm,height=4.0cm]{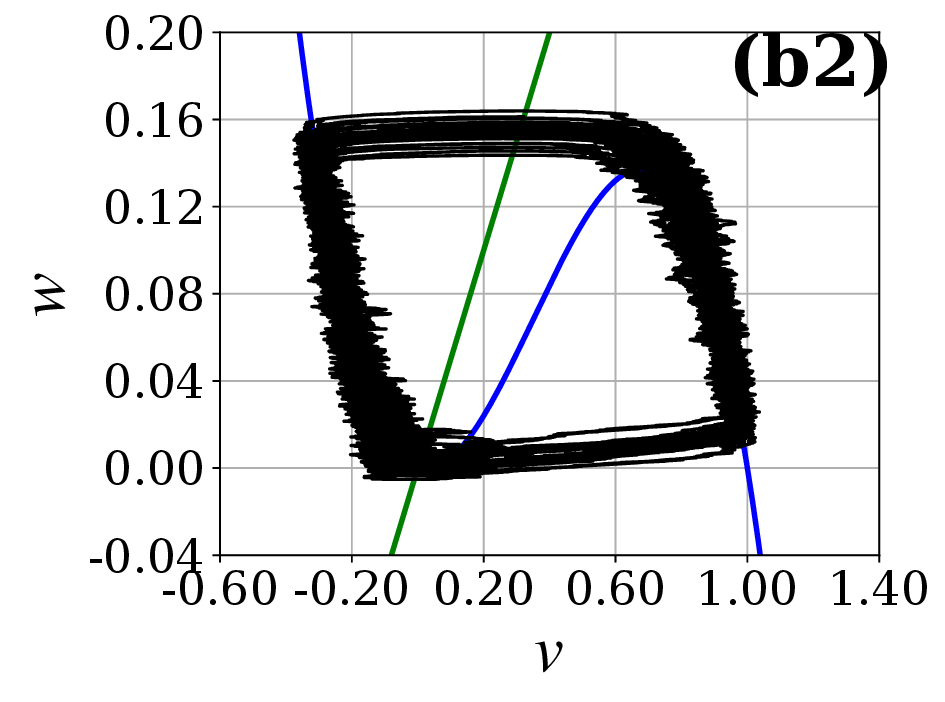}
\includegraphics[width=5.0cm,height=4.0cm]{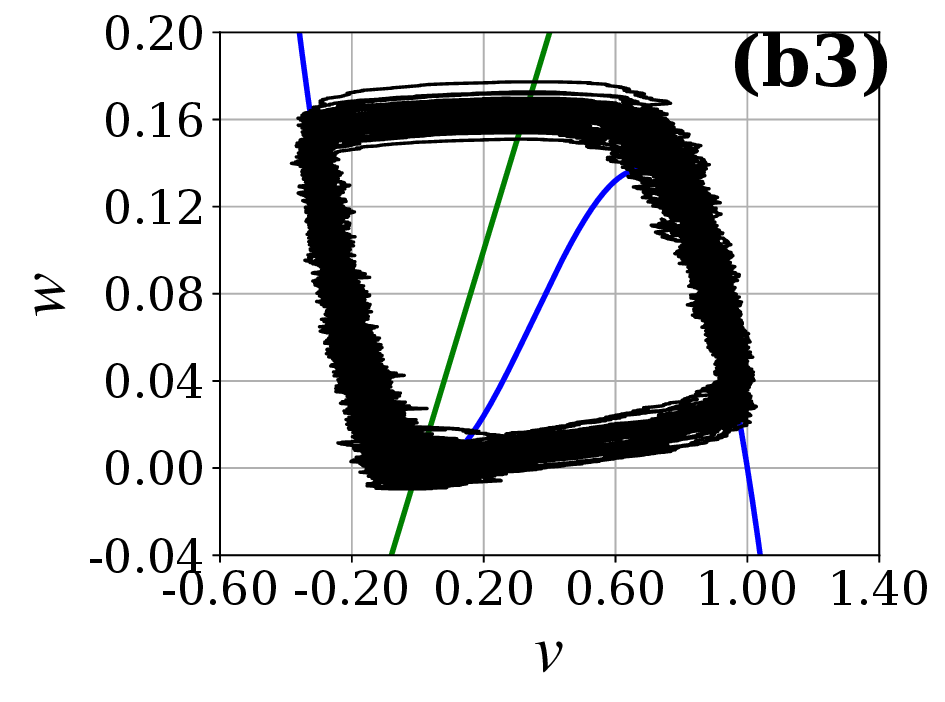}\\
\includegraphics[width=6.0cm,height=4.0cm]{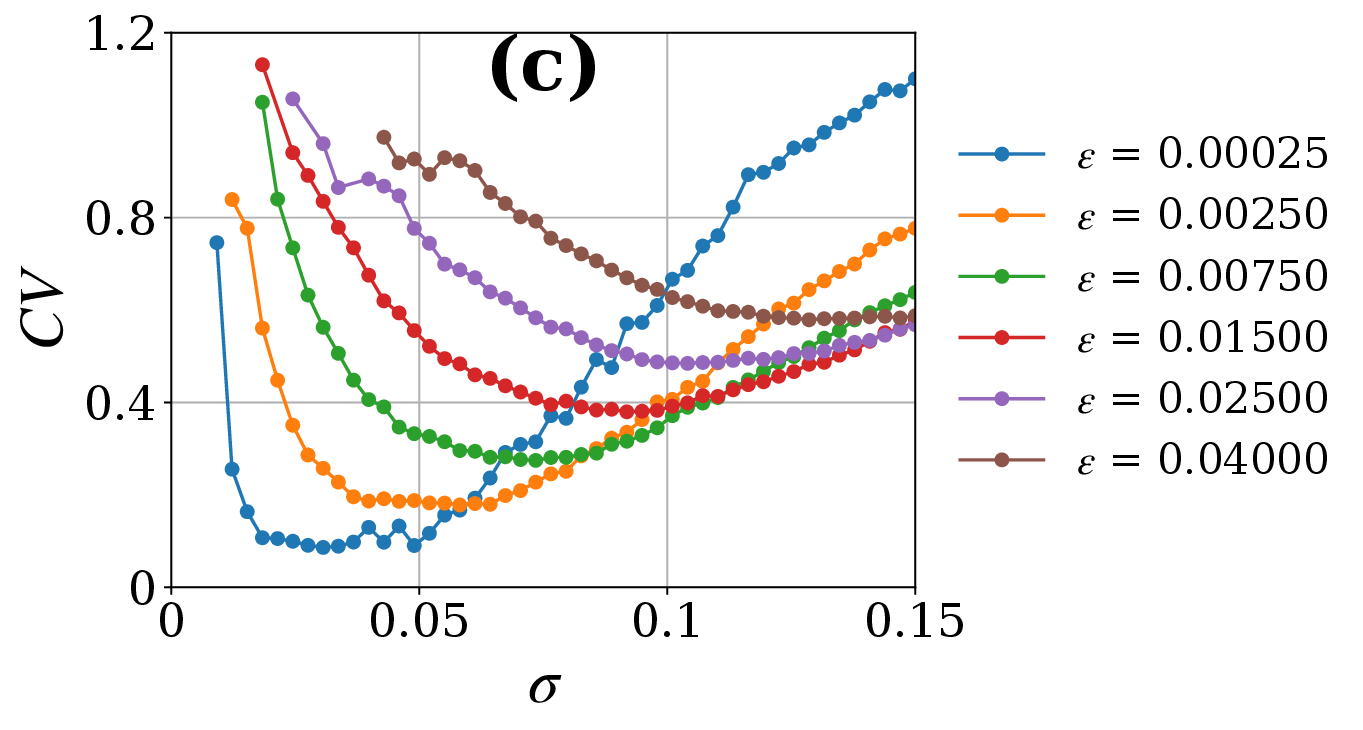}
\caption{Effect of the timescale parameter $\varepsilon$ on SISR. (a1)--(a3) Time series of
$v(t)$ (blue) and $w(t)$ (red) at $\sigma=0.05$. (b1)--(b3) Corresponding phase portraits.
(c) $\mathrm{CV}$ vs. $\sigma$ for different $\varepsilon$. Smaller $\varepsilon$ enhances SISR coherence at lower noise intensities. Parameters: $a=0.05$, $b=1.0$, $c=2.0$.}
\label{fig:5}
\end{figure}

Numerical simulations were performed using the Euler--Maruyama scheme~\cite{highamAlgorithmicIntroductionNumerical2001} with step size $\Delta t=0.05$. Long integration times (see the time series in Figs.~\ref{fig:4} and~\ref{fig:5}) were required to ensure the collection of sufficient ISI samples for statistical convergence of $\mathrm{CV}$. The results are summarized in Figs.~\ref{fig:4} and~\ref{fig:5}.

Figure~\ref{fig:4} shows the influence of $a \in (0, 1+\sqrt{2})$ on SISR at fixed $\varepsilon=0.00025$. Increasing $a$ reduces both the frequency and regularity of noise-induced spikes, as evident in the time series (a1)--(a3) and phase portraits (b1)--(b3). The fixed point, given by the intersection of the $v$- and $w$-nullclines, moves farther from the minimum of the $v$-nullcline as $a$ increases, enhancing excitability (\textit{i.e.,} rendering the stable fixed point more strongly attracting) but degrading temporal coherence. This behavior is quantified in Fig.~\ref{fig:4}(c): smaller $a$ yields lower $\mathrm{CV}$ minima at weaker noise, \marius{whereas} larger $a$ shifts the minima to higher $\sigma$ values, consistent with a loss of timescale matching in Eq.~\eqref{eq:15}.

Figure~\ref{fig:5} explores the effect of $\varepsilon$ at fixed $a=0.05$. Increasing $\varepsilon$ leads to more frequent but less coherent spiking, without changing the fixed point location, as seen in panels (b1)--(b3). This reflects a purely dynamical effect: larger $\varepsilon$ shortens the deterministic timescale $\varepsilon^{-1}$ along the stable nullcline branches, thereby disrupting the matching condition in Eq.~\eqref{eq:15}. As shown in Fig.~\ref{fig:5}(c), $\mathrm{CV}$ minima occur at smaller $\varepsilon$,  \marius {whereas} larger $\varepsilon$ shifts the minima toward higher noise amplitudes and increases overall $\mathrm{CV}$, confirming the degradation of coherence.

\begin{figure}
\centering
\includegraphics[width=7.0cm,height=5.0cm]{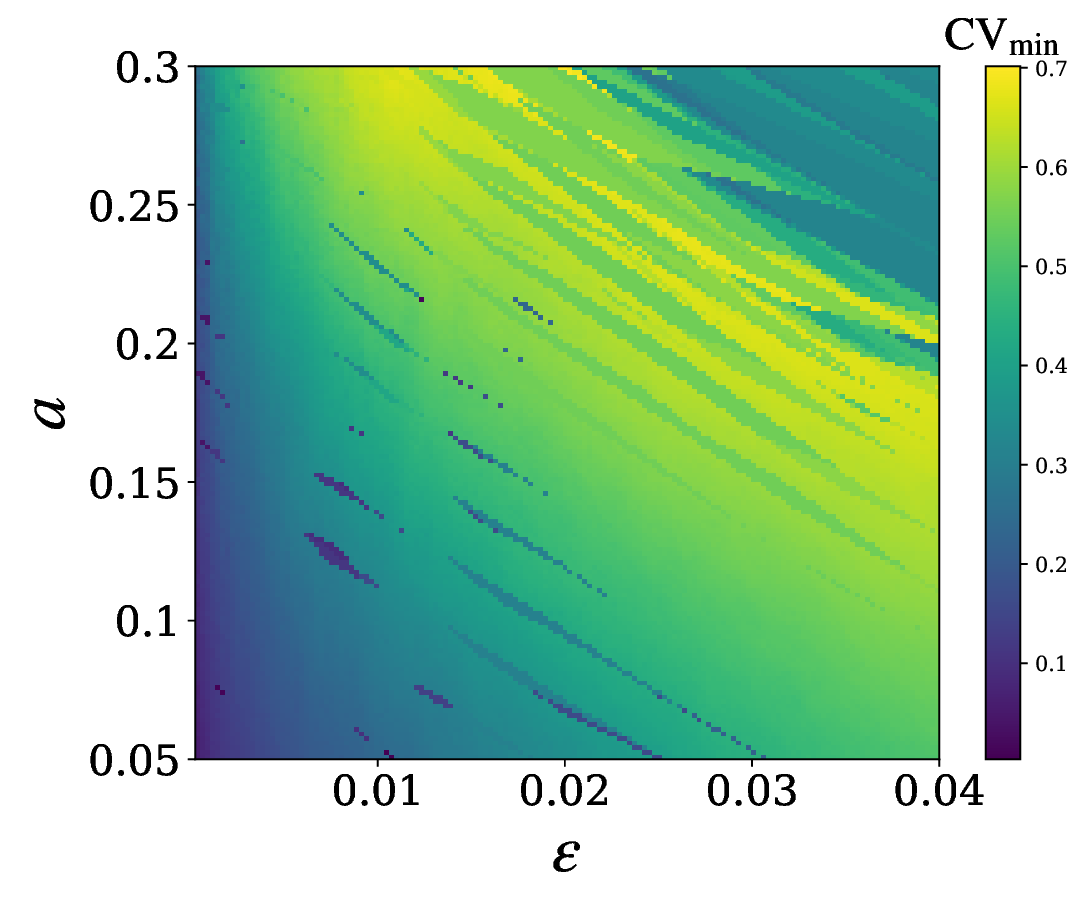}
\caption{Dependence of SISR coherence on $a$ and $\varepsilon$. The color scale represents the minimum coefficient of variation,
$\mathrm{CV}_{\min}$, computed over varying noise intensities for each pair $(a,\varepsilon)$. 
A region of low $\mathrm{CV}_{\min}$ values marks the parameter regime where the 
deterministic relaxation timescale $\varepsilon^{-1}$ and the mean noise-induced escape time 
are optimally matched, resulting in maximal coherence of noise-induced oscillations. $b=1.0$, $c=2.0$.}
\label{fig:6}
\end{figure}

Figure~\ref{fig:6} shows how the coherence of self-induced stochastic resonance depends jointly 
on the excitability parameter~$a$ and the timescale separation~$\varepsilon$. Each color level 
represents the minimum coefficient of variation, $\mathrm{CV}_{\min}$, obtained across noise 
intensities for a given pair $(a,\varepsilon)$. A narrow ridge of low $\mathrm{CV}_{\min}$ values 
marks the region of most coherent spiking, where the deterministic relaxation timescale 
$\varepsilon^{-1}$ and the mean noise-induced escape time are optimally matched. Increasing 
either $a$ or~$\varepsilon$ shifts the system away from this balance, reducing coherence. 
Hence, Fig.~\ref{fig:6} provides a compact overview of the parameter regime where SISR is strongest.

\marius{Although} these simulations validate the theoretical SISR conditions, direct numerical integration is computationally expensive, requiring long simulations to resolve rare stochastic transitions. Moreover, such methods struggle with noisy, sparse, or incomplete data, which is typical of experimental settings. PINNs provide an efficient alternative by directly embedding the governing equations into the learning objective, enabling the data-driven reconstruction of SISR dynamics with limited data and under physical constraints. The next section develops this framework in detail.

\section{SISR: A physics-informed neural network approach}
\label{sec:sisr_Pinn}

Building on the dynamical-systems analysis of Sec.~\ref{SISR_dynamical_system_approach}, we now explore a complementary, data-driven perspective in which SISR is modeled using PINNs. In this framework, PINNs serve not only as computational solvers for stochastic differential equations but also as interpretable models that integrate physical constraints into the learning process. We first recall the multilayer perceptron (MLP) as the underlying architecture and then describe how physical laws are embedded into the network’s loss function, providing a physics-constrained formulation of SISR dynamics.

\subsection{Multilayer perceptron}
\label{sec:mlp}
To emulate stochastic FHN dynamics efficiently, we employ a 
\textit{Noise-Augmented State Predictor (NASP)} as the neural backbone. 
NASP is a data-driven, one-step mapping from the current system state and instantaneous 
noise increment, $(v_t, w_t, \sigma \eta_t)$, to the next-step prediction 
$(\hat{v}_{t+\Delta t}, \hat{w}_{t+\Delta t})$ \cite{hramovForecastingCoherenceResonance2024}. 
By explicitly conditioning on the realized stochastic input $\sigma \eta_t$, 
NASP learns a noise-aware integration rule that captures the impulsive influence of noise 
on the fast subsystem $v$ while preserving stability of the slow variable $w$. 
In this sense, NASP acts as a learned Euler--Maruyama integrator, forming the foundation 
for the subsequent physics-informed extensions \cite{DridiDrumetzFablet2021}.

The multilayer perceptron (MLP) is a fundamental feed-forward neural network architecture that forms the basis of most modern deep learning models~\cite{ramchounMultilayerPerceptron2016,luDeepXDEDeepLearning2021}. It represents a parametric mapping from inputs to outputs through a sequence of affine transformations and nonlinear activations. When multiple hidden layers are stacked, the network acquires substantial expressive power, enabling the approximation of complex nonlinear functions.

The network parameters (weights and biases) are optimized by minimizing a loss function quantifying the mismatch between predictions and data. Gradients are computed via backpropagation, and optimization is typically performed using gradient-based algorithms such as stochastic gradient descent, Adam~\cite{bottouOptimizationMethodsLargeScale2018}, or L-BFGS~\cite{raissiPhysicsInformedNeural2019,markidisOldNewPINNs2021}.

The MLP architecture employed in this work, illustrated in Fig.~\ref{fig:7}, follows a standard fully connected design~\cite{cuomoScientificMachineLearning2022,luDeepXDEDeepLearning2021}. The input layer receives the FHN state variables $(v_t, w_t)$ together with the stochastic input $\sigma \eta_t$, where $\sigma$ denotes the noise amplitude. These quantities are propagated through several hidden layers using hyperbolic tangent activations. The final linear layer outputs the predicted states $(\hat{v}, \hat{w})$. This network serves as the backbone onto which physics-informed constraints are imposed in the next subsection.

\subsection{Physics-informed neural networks}
\label{sec:pinn}
\marius{Although} conventional neural networks can approximate arbitrary nonlinear mappings, their accuracy depends heavily on large amounts of labeled data—often impractical in scientific contexts. PINNs overcome this limitation by embedding the governing differential equations directly into the loss functional~\cite{karniadakisPhysicsInformedMachine2021}. This strategy combines data-driven learning with physics-based regularization, enabling efficient inference from sparse or noisy observations while maintaining consistency with the underlying system dynamics. In this sense, PINNs provide a principled bridge between stochastic dynamical modeling and data-driven learning, allowing the SISR mechanism to be reconstructed and analyzed within a unified framework.

\begin{figure}
    \centering
    \includegraphics[width=21.0cm,height=9.0cm]{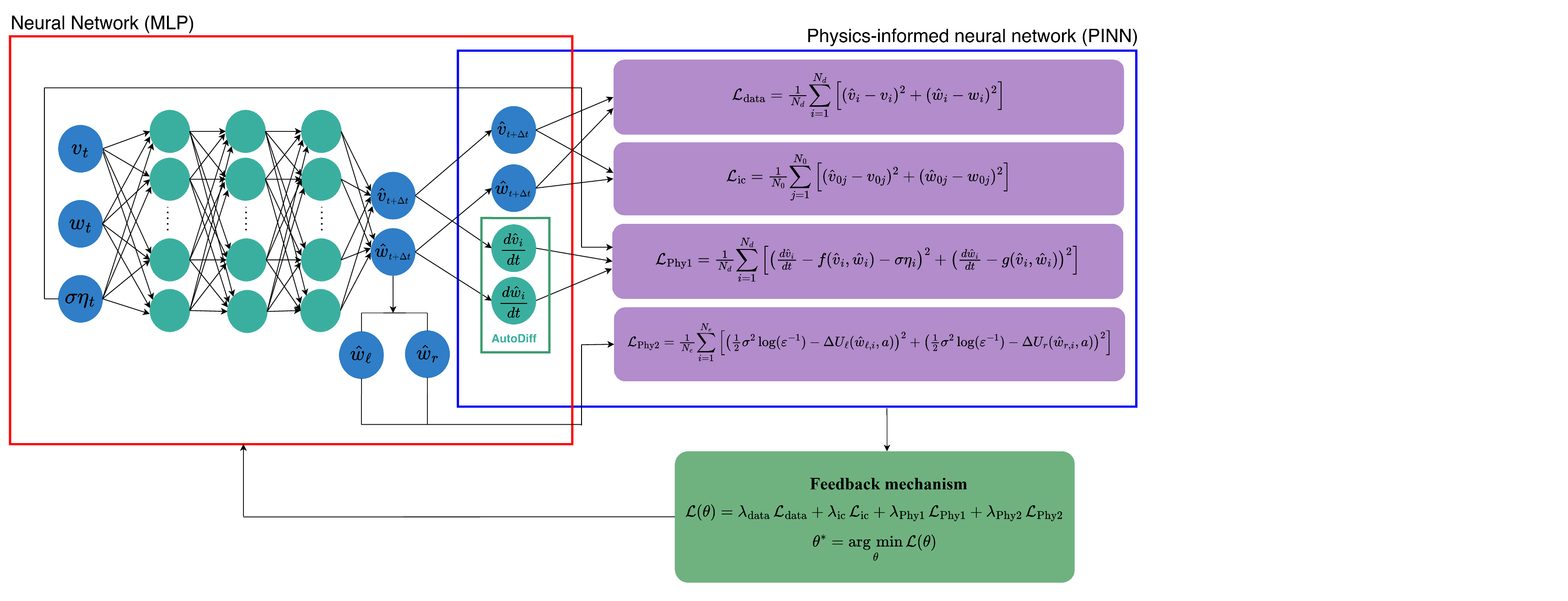}
\caption{PINN architecture $\mathcal{N}(v,w,\sigma\eta,\theta)$ to predict SISR. The backbone is a feed-forward MLP (left) mapping the state 
variables $(v,w)$ and noise $\sigma\eta$ to predicted states $(\hat{v},\hat{w})$. 
The physics-informed extension (right) augments training with a composite loss functional that 
combines data error,  initial-condition penalties, dynamical residuals, and \marius{the barrier-based 
constraints} of SISR in Eq.~\eqref{eq:15}, enforcing consistency between predictions 
and the governing equations.}
\label{fig:7}
\end{figure}

Figure~\ref{fig:7} shows the resulting architecture. The backbone is the feed-forward MLP 
described in Sec.~\ref{sec:mlp}, which maps the inputs $(v_t,w_t,\sigma\eta_t)$ to the predicted states 
$\hat{v}=\hat{v}_{t+\Delta t}$ and $\hat{w}=\hat{w}_{t+\Delta t}$. The physics-informed extension augments training with a composite 
loss $\Lb(\theta)$ that combines supervised data error, the residual of the stochastic dynamics, and penalties for 
deviations from initial conditions, and additional terms enforcing the matching between the deterministic relaxation along the slow manifold and the stochastic escape time across potential barriers, as in Eq. \eqref{eq:PINN_loss_summary}:  
\begin{align}\label{eq:PINN_loss_summary}
\Lb(\theta) &= \lambda_\text{\tiny{data}}\Lb_\text{\tiny{data}}+ \lambda_\text{\tiny{ic}}\Lb_\text{\tiny{ic}}+ \lambda_\text{\tiny{phy1}}\Lb_\text{\tiny{phy1}} + \lambda_\text{\tiny{phy2}}\Lb_\text{\tiny{phy2}}\nonumber \\[1.0mm]
&= \tfrac{\lambda_\text{\tiny{data}}}{N_d}\!\sum_{i=1}^{N_d} 
\Big[(\hat{v}_i - v_i)^2 + (\hat{w}_i - w_i)^2 \Big] +  \tfrac{\lambda_\text{\tiny{ic}}}{N_0}\!\sum_{j=1}^{N_0} 
\Big[  (\hat{v}_{0j}-v_{0j})^2 + (\hat{w}_{0j}-w_{0j})^2 \Big] \nonumber \\[1.0mm]
&+\tfrac{\lambda_\text{\tiny{phy1}}}{N_d}\!\sum_{i=1}^{N_d} 
\Big[ \Big(\frac{d\hat{v}_i}{dt} - f(\hat{v}_i,\hat{w}_i)-\sigma\eta_i\Big)^2 + \Big(\frac{d\hat{w}_i}{dt}  - g(\hat{v}_i,\hat{w}_i)\Big)^2 \Big] \nonumber \\[1.0mm]
&+
\tfrac{\lambda_\text{\tiny{phy2}}}{N_e}\!\sum_{i=1}^{N_e} 
\Big[\Big(\frac{1}{2}\sigma^2\log(\varepsilon^{-1}) - \Delta U_{\ell}(\hat{w}_{\ell,i},a)\Big)^2 + \Big(\frac{1}{2}\sigma^2\log(\varepsilon^{-1}) - \Delta U_{r}(\hat{w}_{r,i},a)\Big)^2 \Big].
\end{align}
The supervised data term enforces agreement between the network predictions and available data via
$\lambda_\text{\tiny{data}}\Lb_\text{\tiny{data}}$, where $\{(v_i,w_i,\sigma\eta_i)\}_{i=1}^{N_d}$ are training samples and $(\hat{v}_i,\hat{w}_i)$ are the corresponding network one-time step predicted outputs.

The initial state is enforced by $\lambda_\text{\tiny{ic}}\Lb_\text{\tiny{ic}}$
where $\{(v_{0j},w_{0j})\}_{j=1}^{N_0}$ specify the prescribed initial conditions.
In our setting, we have $N_0=1$, so the sum reduces to a single term, but the general 
formulation is kept for consistency with the other loss components. 

The residual loss enforces consistency with the stochastic FHN dynamics via $\lambda_\text{\tiny{phy1}}\Lb_\text{\tiny{phy1}}$ where $f$ and $g$ denote the model deterministic vector field, and predicted derivatives denoted by $(d\hat{v}_i/dt,d\hat{w}_i/dt)$ are obtained from the network by automatic differentiation (AutoDiff) \cite{raissiPhysicsInformedNeural2019}.

Finally, the barrier-based physics loss $\lambda_\text{\tiny{phy2}} \Lb_{\text{\tiny{phy2}}}$ enforces the 
asymptotic matching of deterministic and stochastic timescales, as required by the third 
condition for SISR in Eq.~\eqref{eq:15}. Here, 
$\Delta U_{\ell}(\hat{w}_{\ell,i},a)$ and $\Delta U_{r}(\hat{w}_{r,i},a)$ denote the left and right 
network-predicted barrier terms, evaluated at the predicted escape points 
$\hat{w}_{\ell,i}$ and $\hat{w}_{r,i}$ for $i=1,2,\ldots,N_e$. The integer $N_e$ corresponds 
to the number of escape events of the predicted trajectory $\hat{w}(t)$ from the left 
($\hat{w}_{\ell}$) and right ($\hat{w}_{r}$) branches of the $\hat{v}$-nullcline.
Over a given time interval, the numbers of predicted escape points from the left (\(\hat{w}_{\ell,i}\)) and right (\(\hat{w}_{r,i}\)) sides are not necessarily equal for a predicted trajectory \(\hat{w}(t)\). Consequently, the number of escape events \(N_e\) may differ between the two directions, in which case the loss \(\Lb_{\text{\tiny{phy2}}}\) is computed using the corresponding event counts.

The $\lambda_{\tiny{data}}, \lambda_{\tiny{ic}}, \lambda_{\tiny{phy1}}, 
\lambda_{\tiny{phy2}}\geq 0$ are tunable hyperparameters that control the relative 
importance of each loss component in the 
loss functional $\Lb(\theta)$  during training. In practice, these weights can be 
chosen empirically, adapted dynamically, or normalized to balance gradient magnitudes 
across the different terms \cite{mcclennySelfAdaptivePhysicsInformed2023, wangWhenWhyPINNs2022}. In our case, we adapt them dynamically during training. In the next section, we describe the preparation of the training data, the setup of the model and optimizer, and the results of training and validating the PINN for SISR prediction.

It is worth noting that our goal is not merely to numerically solve the stochastic FHN model---which can already be done using a standard Euler-Maruyama integrator---but rather to learn
a data-driven surrogate that reproduces both short-term dynamics and long-term
stochastic statistics of SISR from relatively short simulated trajectories. The NASP–PINN learns the stochastic one-step transition map $(v_t, w_t, \sigma\eta_t) \mapsto (v_{t+\Delta t},
w_{t+\Delta t})$ and can subsequently generate long rollouts, ISI statistics, and $\mathrm{CV}$-noise
curves without reintegrating the SDE, acting as a learned Euler-Maruyama-type
integrator with reduced computational cost compared to Monte Carlo simulation.

The physics losses serve as inductive priors rather than equation replacements: phy1 enforces
consistency with the deterministic vector field, while phy2 implements the Kramers
timescale-matching mechanism underlying SISR, jointly improving training and generalization. 
This work, therefore, operates in the
\textit{physics-known regime}, where the full FHN equations provide a controlled environment
to demonstrate the benefit of physically guided learning. In more realistic scenarios with unknown or
incomplete physical knowledge (\textit{e.g.,} experimental recordings), phy1/phy2 could be replaced
or complemented by alternative priors such as data-driven drift/diffusion estimation,
nullcline inference, escape-time constraints, SINDy-based sparse regression \cite{brunton2016discovering,lu2022discovering}, or other
structural properties. This study thus establishes a benchmark for physics-regularized
surrogate modeling of SISR and motivates future extensions to partially known systems.

\subsection{Training data preparation}
\label{sec:training_data}

A defining feature of SISR is its emergence in the singularly perturbed limit 
$\varepsilon \to 0$, where the deterministic relaxation timescale scales as 
$\varepsilon^{-1}$ and becomes large. The resulting slow drift along the 
$v$--nullcline enables timescale matching with stochastic escapes, but also 
introduces a practical challenge: accurately estimating spike-train coherence 
(\textit{e.g.,} via the coefficient of variation, $\mathrm{CV}$) requires a sufficiently 
large ensemble of interspike intervals (ISIs), which demands long numerical 
integrations. As discussed earlier, $\mathrm{CV}$ only becomes meaningful once 
at least $3$ ISIs are available; however, reliable estimation generally requires 
significantly more samples ($\geq 10$ ISIs depending on variability). In practice, 
collecting even $5$--$10$ ISIs in the adiabatic regime $\varepsilon \to 0$ may 
require integration over $(15\times10^{3}$--$25\times10^{3})$ time units, 
since spike events become increasingly rare as $\varepsilon$ decreases.

This computational bottleneck motivates the use of our NASP--PINN. Rather than 
performing prohibitively long simulations, we train the network using comparatively 
short trajectories of $T = 10\times10^{3}$ time units ($2\times10^{5}$ steps with 
$dt = 0.05$), which contain only $3$ ISIs for $\varepsilon = 0.00025$. Such a window 
is insufficient for a reliable $\mathrm{CV}$ estimate, but provides enough local 
dynamical variability for NASP--PINN to learn the stochastic transition law. Once 
trained, the model is rolled out autonomously to generate long-horizon trajectories 
from which reliable $\mathrm{CV}$ can be estimated efficiently and without further numerical 
integration.

Unless stated otherwise, the training data consist of simulated trajectories 
$(v_t, w_t, \eta_t)$ generated by numerically integrating Eq.~\eqref{eq:1} over 
$t \in [0, 1.0\times10^{4}]$ with time step $\Delta t = 0.05$, yielding 
$N_d = 2\times10^{5}$ data points. Parameters are $a = 0.05$, $b = 1.0$, 
$c = 2.0$, $\varepsilon = 0.00025$, and $\sigma = 0.03061$.

\subsection{Model, optimizer setup, and hyperparameters}
\label{sec:optimization}
Training is performed via backpropagation using the Adam optimizer, see the algorithm in the Appendix \ref{appendix}. We initialize a StatePredictor with three 128-unit Tanh blocks and a dynamics head and set $dt=0.05$, epochs $=1\times10^{4}$, batch size $=512$, and learning rate $=1.0\times10^{-3}$. The total loss is a weighted sum of the data loss, the initial-condition loss, the residual loss, and the barrier-based physics loss weighted by, $\lambda_{\tiny{data}}, \lambda_{\tiny{ic}}, \lambda_{\tiny{phy1}}, \lambda_{\tiny{phy2}}$ respectively. During training, each minibatch is a consecutive time sequence. Because the loss weights $\lambda_i$ are dynamically adjusted during training, we additionally report how much
each loss component contributes to the parameter updates over time. For each component $\Lb_{i}(\theta)$, we compute
the gradient-weighted share
\begin{equation}\label{eq:gradient}
s_i(\text{epoch}) \;=\; \frac{\lambda_i \,\lVert \nabla_{\theta} \Lb_{i} \rVert}{\sum_{j} \lambda_j \,\lVert \nabla_{\theta} \Lb_{j} \rVert},
\end{equation}
so that $\sum_i s_i = 1$ at every epoch, where $i, j\in\{\text{data, ic, phy1, phy2}\}$. This diagnostic reveals which term dominates the optimization at
different phases of learning  \cite{chenGradNormGradient2018}.

For evaluation, we predict the target sequence and compute the normalized root-mean-square error (NRMSE).
The optimal training epoch is selected as the one yielding the smallest NRMSE, defined for a reference sequence \(y = \{y_t\}_{t=1}^N\) and its predictions \(\hat{y} = \{\hat{y}_t\}_{t=1}^N\) by
\begin{equation}\label{eq:nrmse}
\mathrm{NRMSE}(y, \hat{y}) =
\frac{\sqrt{\tfrac{1}{N}\!\sum_{t=1}^{N} (y_t - \hat{y}_t)^2}}
{\sqrt{\tfrac{1}{N}\!\sum_{t=1}^{N} (y_t - \bar{y})^2}},
\end{equation}
Here $y = (v(t), w(t))$ denotes the two-dimensional reference trajectory of the 
system state, and $N$ is the number of time points in the evaluated prediction 
window. The NRMSE therefore aggregates errors across both the fast variable $v$ 
and the slow variable $w$, reflecting the need for the surrogate to jointly 
predict both components of the slow--fast dynamics. Lower NRMSE values indicate 
better predictive performance; consequently, a faster decrease in NRMSE during 
training corresponds to faster convergence of the optimization process.

It is worth mentioning that several training experiments (not shown) were performed using NumPy and PyTorch with different initialization seeds and all showed the same qualitative convergence and rollout behavior, confirming that the NASP--PINN is not sensitive to different initializations.

\subsection{Results on training and evaluation}
We evaluate the NASP--PINN on held-out sequences to measure both one-step 
prediction quality and multi-step rollout fidelity. In addition, we conduct a 
formal ablation study to quantify the individual contribution of each 
physics-informed component of the loss. Specifically, we benchmark four model 
variants: (i) data only, (ii) data + ic + phy1 (SDE residual loss), 
(iii) data + phy2 (Kramers barrier-matching), and (iv) the full model with all 
terms included. This ablation design isolates the effect of each physics term on 
optimization behavior, predictive accuracy, and out-of-sample performance.

For each experiment, the stochastic FHN system is integrated over a fixed time 
horizon $t \in [0, 25\times10^{3}]$, which produces $5\times10^{5}$ data points 
with $\Delta t = 0.05$. The training window corresponds to the initial segment 
$t \in [0, 10\times10^{3}]$, processed in consecutive minibatches as described 
in Section~\ref{sec:optimization}. The held-out test sequence refers to 
the immediately following portion of the same simulated trajectory, 
\textit{i.e.,} $t \in (10\times10^{3}, 25\times10^{3}]$, which is not used during 
training and therefore serves as unseen future dynamics for evaluating temporal 
generalization. The lengths of the training and testing windows used in all 
quantitative evaluations are identical.

To compute the NRMSE (Eq.~\eqref{eq:nrmse}), a ground-truth reference trajectory 
of the same length as the model prediction is required. We therefore designate the trajectory obtained by numerically integrating 
Eq.~\eqref{eq:1} as the \emph{reference sequence}, and evaluate all model 
variants against this same trajectory.
 All model variants are evaluated against this 
same reference, ensuring a consistent and fair comparison across ablation 
settings. This reference sequence is representative because it is drawn from the 
stationary stochastic dynamics, and SISR statistics (such as the ISI 
distribution and the slow drift of $w$) remain stable for fixed parameter values used.

\begin{figure}
\centering
\includegraphics[width=8.0cm,height=5.0cm]{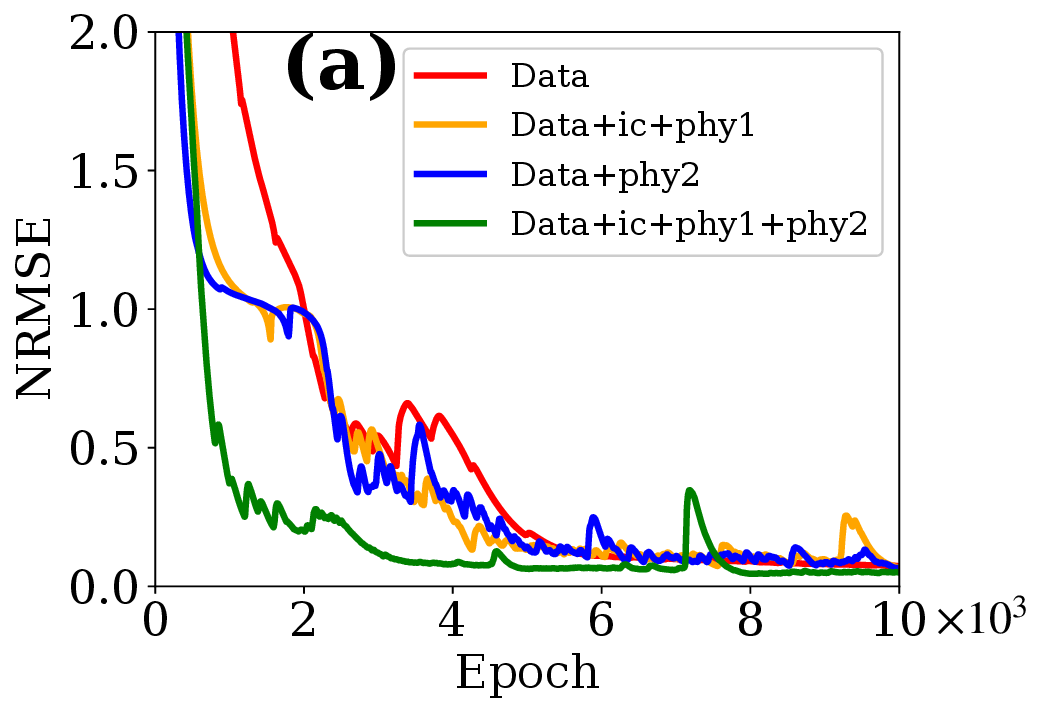}\includegraphics[width=8.0cm,height=5.0cm]{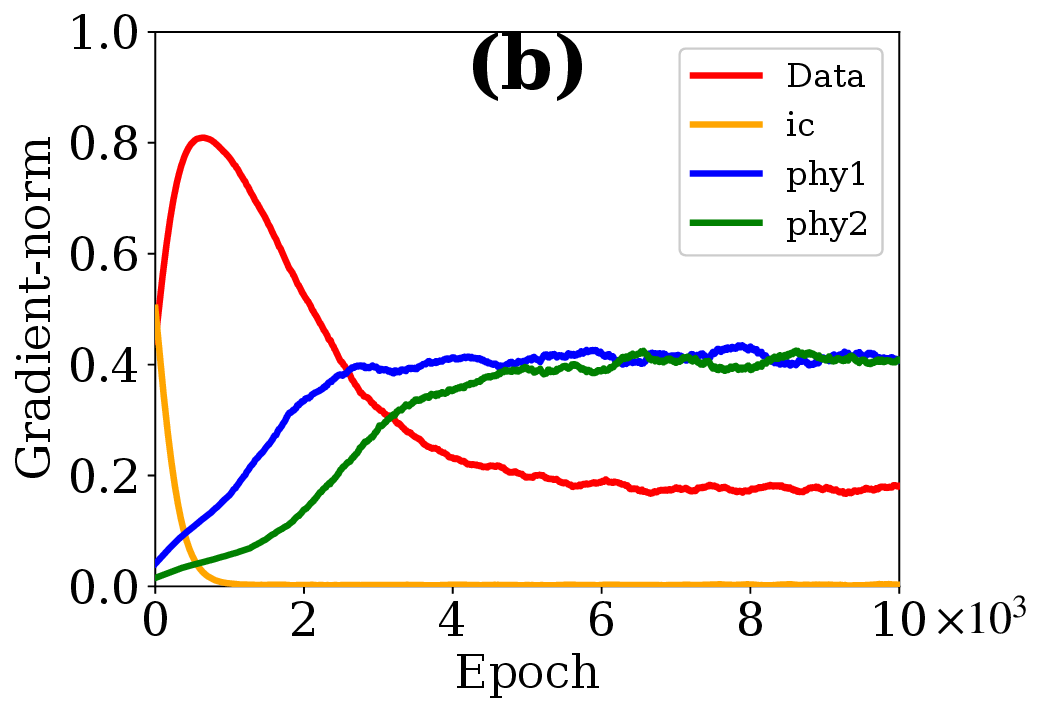}
\caption{(a) NRMSE values during training. Ablation study of NASP--PINN training performance under different loss  configurations. Shown are NRMSE curves for four variants: data only (red), 
data+ic+phy1 (yellow), data+phy2 (blue), and data+ic+phy1+phy2 (green). The 
physics components (phy1, phy2) significantly accelerate convergence and reduce 
error. The full model exhibits the best optimization behavior, achieving the 
fastest convergence, most stable training, and the best test performance 
(see Table~\ref{tab:1}). (b) Effective gradient contribution of each loss component for the full model, computed as $s_i$ (Eq.~\eqref{eq:gradient}). Shows how the optimization focus transitions across training, indicating which constraints dominate parameter updates over time.}
\label{fig:8}
\end{figure}

\begin{figure}
\centering
\includegraphics[width=7.0cm,height=4.0cm]{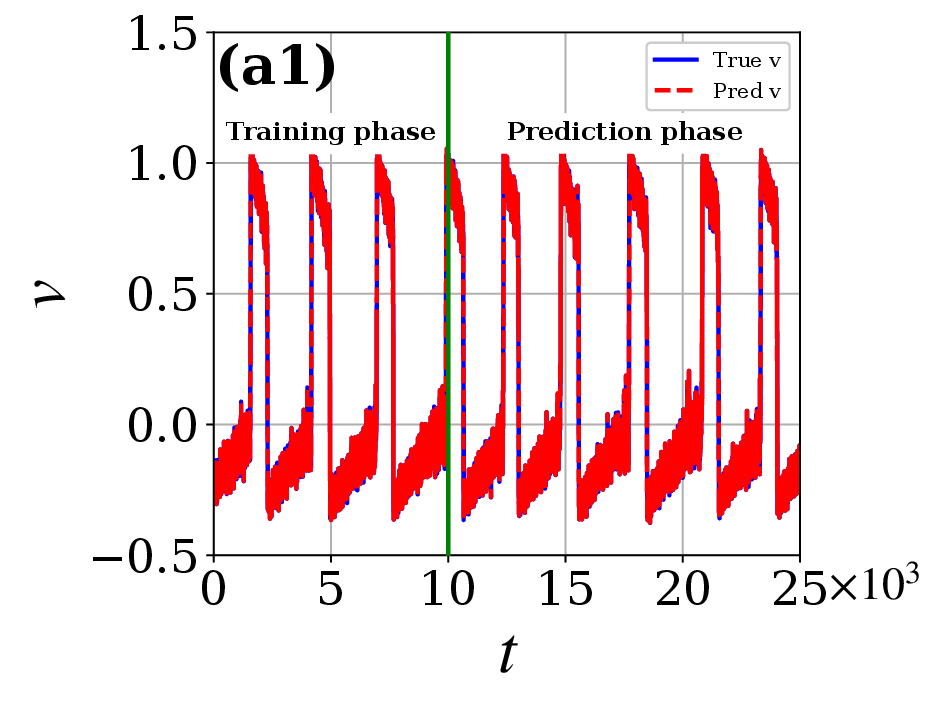}
\includegraphics[width=7.0cm,height=4.0cm]{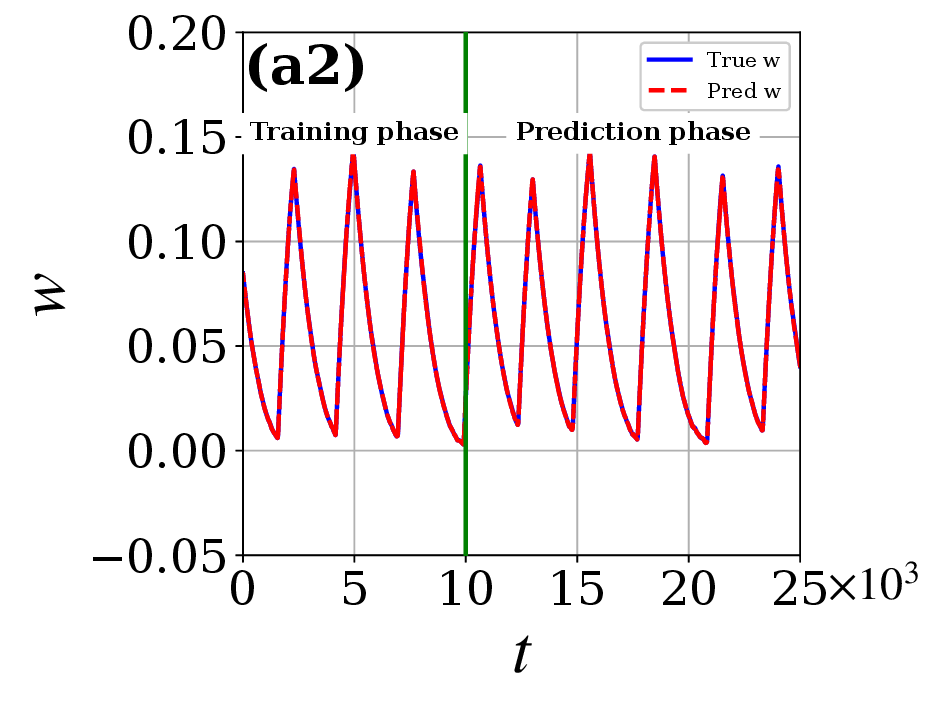}\\
\includegraphics[width=7.0cm,height=4.0cm]{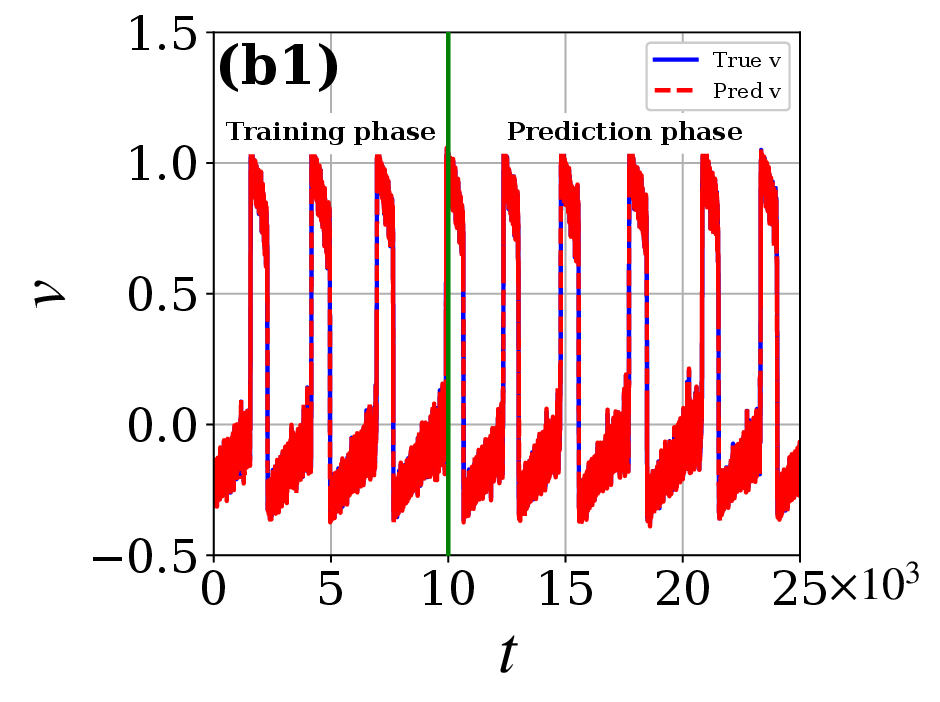}
\includegraphics[width=7.0cm,height=4.0cm]{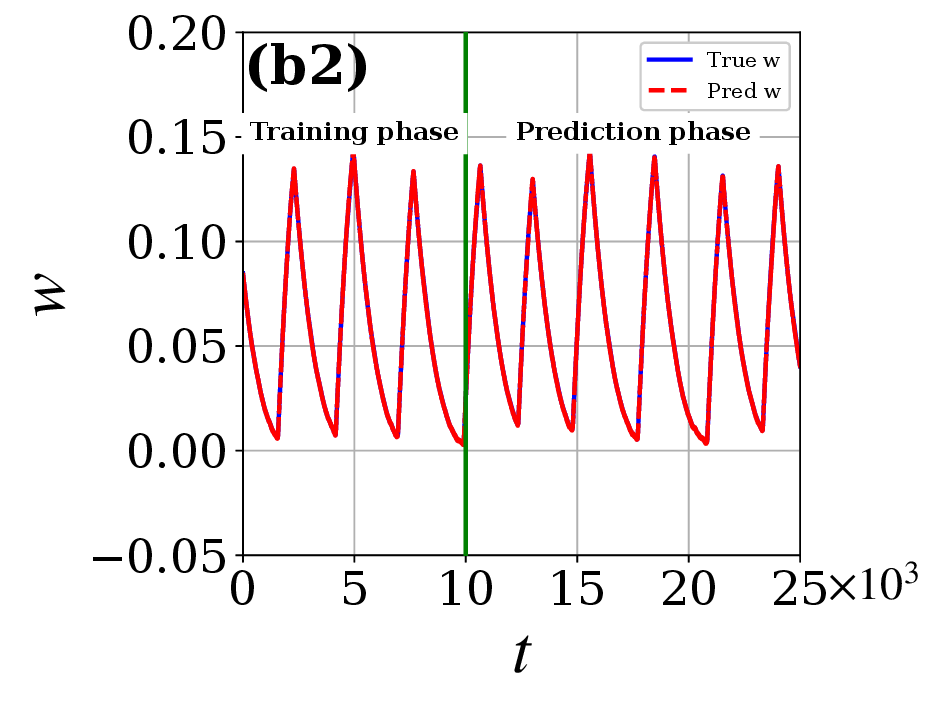}
\caption{
Comparison of rollout performance for the data-only model (top row) and the 
full NASP--PINN (bottom row). Panels (a1)--(a2) show the predicted and true 
time series of $v(t)$ and $w(t)$ for the data-only model. Panels 
(b1)--(b2) show the corresponding results for the physics-informed model. 
Although the trajectories appear visually similar over short horizons, the 
NASP--PINN achieves a substantially lower test NRMSE ($0.035$ vs.\ $0.062$), a 
$43\%$ improvement. In slow--fast excitable systems, such errors accumulate over time and lead to 
shifts in spike timing that directly affect SISR coherence.
}
\label{fig:9}
\end{figure}

The results of this training and evaluation procedure, including the ablation 
study, are summarized in Fig.~\ref{fig:8} and Table~\ref{tab:1}. 
Table~\ref{tab:1} reports the train and test NRMSE for the training window 
$t \in [0,10\times10^{3}]$ and the held-out test window 
$t \in (10\times10^{3},25\times10^{3}]$, using the same reference sequence and 
evaluation protocol for all ablation variants. Figure~\ref{fig:9} illustrates a 
representative free-rollout under this same train/test setup, providing a 
qualitative view of long-horizon behavior—including spike timing, phase 
stability, and amplitude preservation—that complements the quantitative 
performance reported in Table~\ref{tab:1}.

Figure~\ref{fig:8}(a) presents the results of the ablation study using the 
training NRMSE, computed over both state variables $(v,w)$ to capture errors in 
both the fast spiking dynamics and the slow recovery drift. We report the NRMSE 
during training because it provides a physically interpretable and comparable 
measure of prediction quality across the different loss configurations, thereby 
highlighting differences in optimization speed and stability. The four curves 
correspond to the data-only model (red), data+ic+phy1 (yellow), data+phy2 (blue), 
and the full data+ic+phy1+phy2 model (green). Incorporating either of the 
physics-informed terms (phy1 or phy2) accelerates convergence and lowers the 
asymptotic error relative to the purely data-driven baseline. The full NASP--PINN 
achieves the fastest and most stable convergence as well as the lowest NRMSE, 
consistent with the quantitative train/test results reported in 
Table~\ref{tab:1}.

Figure~\ref{fig:8}(b) illustrates how the dynamically weighted objective is optimized over time. Early epochs are dominated by the supervised data and initial-condition terms, reflecting rapid alignment to observed one-step targets. As training progresses, the optimization emphasis shifts toward the physics constraints, with phy1 and phy2 accounting for the majority of the gradient budget in later epochs. This indicates that the physics terms are not negligible and they actively shape the learned dynamics by enforcing local SDE consistency (phy1) and global timescale structure (phy2), which is consistent with the lower test NRMSE of the full model reported in Table~\ref{tab:1}.

Table~\ref{tab:1} summarizes train/test NRMSE for the same variants. Data-only 
training yields the weakest train/test performance (0.053 train, 0.062 test), 
indicating mild overfitting to short-horizon local patterns. Enforcing 
\emph{dynamical residuals} (phy1) and/or \emph{barrier-matching} (phy2) improves 
test performance and lowers both errors (0.040--0.042 train, 0.050 test). The 
\emph{full loss} achieves the best results by a clear margin (0.027 train, 
0.035 test), demonstrating that the two physics priors are complementary:  
phy1 aligns local derivatives with the SDE, while phy2 shapes global exit 
statistics via timescale matching, together constraining the hypothesis class to 
trajectories that are both locally consistent and globally coherent. Relative to 
data-only training, the PINN reduces test NRMSE by approximately 43\% 
(0.062~$\rightarrow$~0.035) while also quickly lowering train error, consistent with a 
better-conditioned objective and stronger inductive bias.
 
\begin{table}
  \centering
  \caption{NRMSE for NASP variants under different loss combinations.}
  \begin{tabular}{lcc}
    \toprule
    Loss Combination         & NRMSE (Train) & NRMSE (Test) \\
    \midrule
    Data                     & 0.053         & 0.062        \\
    Data + ic + phy1         & 0.042         & 0.050        \\
    Data + phy2              & 0.040         & 0.050        \\
    Data + ic + phy1 + phy2  & 0.027         & 0.035        \\
    \bottomrule
  \end{tabular}
  \label{tab:1}
\end{table}

From a learning-theoretic perspective, the physics terms act as inductive biases and implicit regularizers. 
phy1 reduces variance by penalizing derivative mismatches estimated via automatic differentiation; 
phy2 reduces bias in the noise-triggered event timing by anchoring the learned dynamics to the Arrhenius-type scaling required for SISR (timescale matching).  Their combination explains the lower test error and the smoother, faster descent curves in Fig.~\ref{fig:8}(a).  
In broader machine-learning terms, these physics-based penalties constrain the hypothesis space in a way analogous to structural regularization in deep sequence models, improving both data efficiency and temporal consistency.

To test stability beyond the training horizon, we perform open-loop (free-rollout) simulations and compare against direct stochastic simulations. Figure~\ref{fig:9} compares the time series of $v(t)$ and $w(t)$ for the 
data-only model (panels (a1)--(a2)) and the full NASP--PINN 
(panels (b1)--(b2)) across both the training and prediction windows. 
Over the horizon shown, the two models can appear visually similar, as both track 
the large-amplitude spikes in $v$ and the slow drift in $w$. However, the full 
NASP--PINN achieves a substantially lower test NRMSE ($0.035$ vs.\ $0.062$ for 
data-only, see Table \ref{tab:1}), representing a $43\%$ improvement in predictive accuracy. In 
slow--fast excitable systems, such differences can be highly consequential: even 
small local errors accumulate along the slow manifold and lead to progressively 
shifting spike times during very long rollouts. Because SISR coherence depends 
sensitively on the regularity and alignment of these escape events (precision of the spiking times), the 
physics-informed model would definitively yield markedly more reliable very long-horizon dynamics.

The qualitative agreement in panels (b1)--(b2) demonstrates that the learned 
integrator is both \emph{noise-aware} and \emph{stiffness-compatible}, 
properties that are crucial for accurately resolving slow--fast SDEs. The 
barrier-matching term (phy2) teaches the model where and at what rate escapes occur in $w$, \marius{whereas} the residual term (phy1) enforces local dynamical consistency between successive steps. Together, these constraints produce a 
surrogate that maintains phase alignment and amplitude fidelity far more 
robustly than the data-only model, despite the apparent visual similarity over 
relatively short windows.

Overall, the findings demonstrate that embedding the governing stochastic dynamics and \marius{the} SISR-specific constraints directly into the loss function transforms NASP into a data-efficient and \emph{generalizable} framework for modeling noise-driven multiscale dynamics. Across all evaluation metrics—including learning curves, final NRMSE, and long-horizon rollouts—the Data + ic + phy1 + phy2 configuration consistently yields superior performance, with the most pronounced improvements observed in extended predictions where preserving the intrinsic slow–fast structure and accurate escape timing is critical for reproducing SISR. Physics-informed training, particularly the combination of local residual constraints and global barrier-matching, thus serves as an effective inductive bias that enhances generalization while ensuring the model faithfully respects the physics that drives self-induced stochastic resonance.

\begin{figure}
\begin{center}
\includegraphics[width=7.5cm,height=5.0cm]{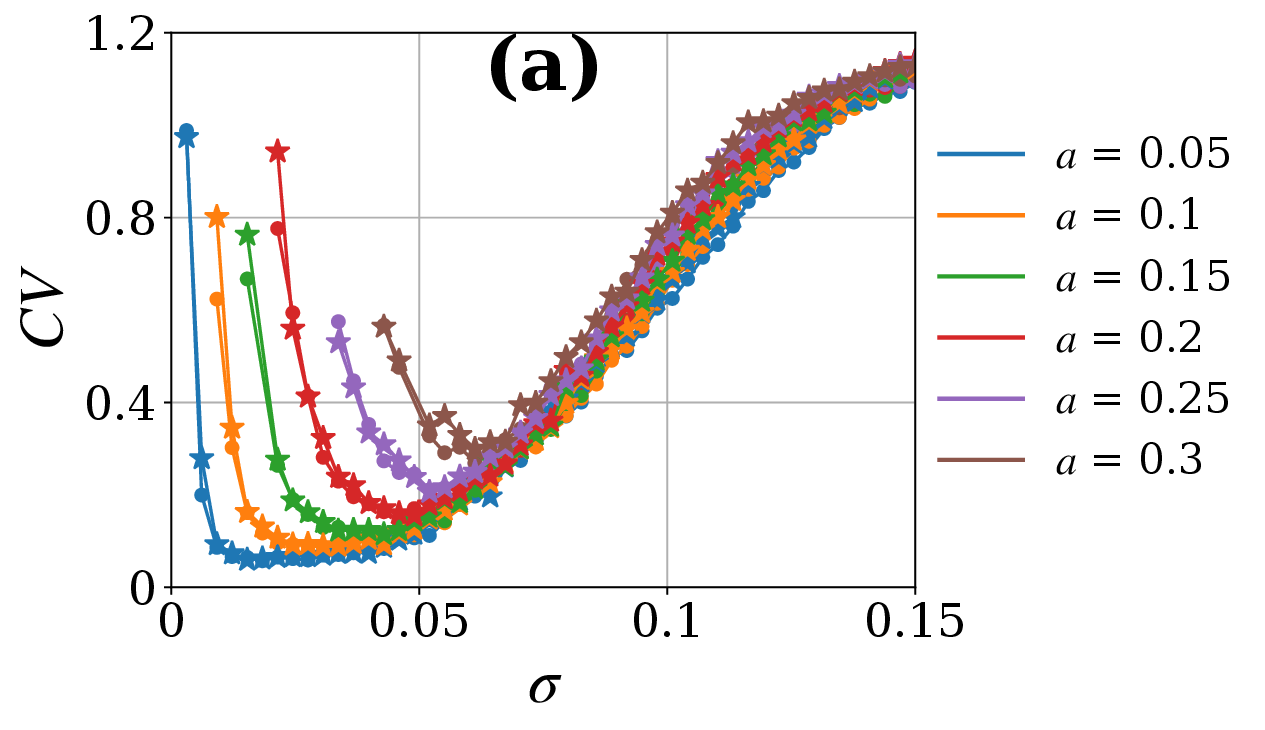}
\includegraphics[width=7.5cm,height=5.0cm]{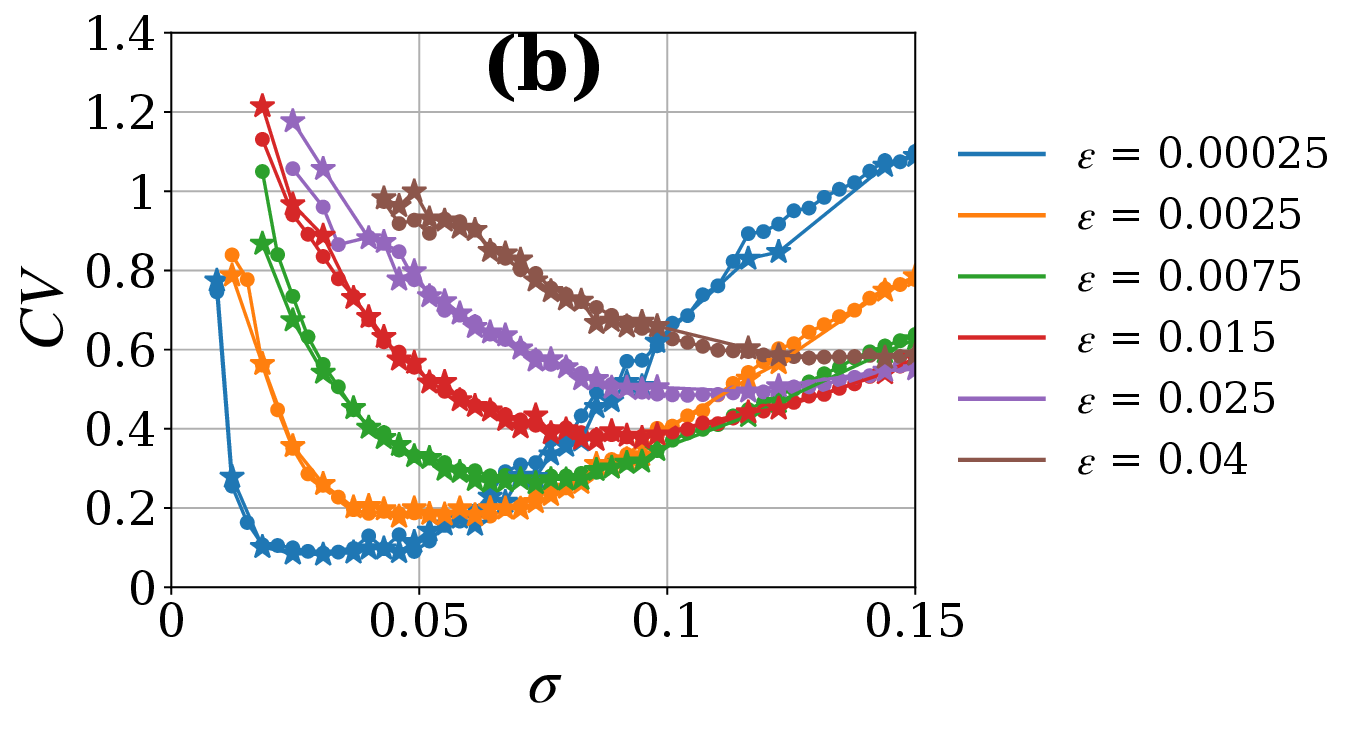}
\end{center}
\caption{True ($\bullet$ marker) and \marius{the} NASP-PINN-predicted ($\star$ marker)  coherence degree of SISR. (a) and (b) show  $\mathrm{CV}$ versus noise intensity $\sigma$ for varying excitability $a$ and timescale separation $\varepsilon$ parameters, respectively, demonstrating the ability of \marius{the} NASP-PINN to predict with a high degree \marius{of} accuracy the dependence of SISR on both parameters. Other parameters: $b=1.0$, $c=2.0$.}
\label{fig:10}
\end{figure}

Figure \ref{fig:10}(a)-(b) compares the noise–coherence relationship predicted by the NASP-PINN with that obtained from direct stochastic simulations of the FHN model Eq. \eqref{eq:1}. The $\mathrm{CV}$ of interspike intervals is plotted as a function of noise intensity $\sigma$ for different values of the excitability parameter $a$ and the timescale separation $\varepsilon$. Both panels show that \marius{the} NASP-PINN does well in predicting the degree of coherence due SISR, as $\sigma$, $a$, and $\varepsilon$ are varied.

The close quantitative agreement between both sets of curves demonstrates that the PINN accurately learns the underlying stochastic dynamics and correctly captures how the coherence of SISR depends on excitability and timescale separation. In particular, the network reproduces the characteristic shift of the $\mathrm{CV}$ minima toward higher noise intensities as $a$ or $\varepsilon$ increase, reflecting the loss of timescale matching between deterministic relaxation and stochastic escape predicted by the theoretical SISR condition. This confirms that embedding the governing physics directly into the training objective enables the PINN to generalize beyond the training data and faithfully recover the resonance structure of the stochastic FHN system.

\section{Summary and conclusions}
\label{sec:conclusion}

In this work, we developed a physics-informed machine learning framework for modeling and predicting self-induced stochastic resonance (SISR) in an excitable FitzHugh--Nagumo (FHN) neuron. We first revisited the phenomenon of SISR from a dynamical-systems perspective, emphasizing its distinct mechanism compared with classical stochastic and coherence resonance. Through analytical and numerical analysis, we showed that SISR emerges from the interplay between deterministic slow--fast dynamics and stochastic escape processes, and that coherence arises when the deterministic relaxation timescale matches the mean noise-induced escape time predicted by Kramers' law.  

We systematically investigated how the excitability parameter~($a$) and the timescale separation~($\varepsilon$) influence the coherence of SISR by computing the coefficient of variation~($\mathrm{CV}$) of interspike intervals from direct stochastic simulations. The results revealed a narrow region in the~($a$,~$\varepsilon$) parameter space where the $\mathrm{CV}$ reaches its minimum, marking the optimal balance between deterministic and stochastic timescales and thus maximal coherence of noise-induced oscillations.  

Building on these theoretical insights, we introduced a Physics-Informed Neural 
Network (PINN) based on a Noise-Augmented State Predictor (NASP) architecture to 
learn the stochastic FHN dynamics directly from data. The composite loss 
integrates data fidelity, dynamical residuals, and \marius{the} physics-based constraints 
derived from SISR theory. Through a dedicated ablation study, we quantified the 
individual contributions of each loss component—data term, SDE residual (phy1), 
and Kramers timescale-matching constraint (phy2)—and observed that each provides 
a meaningful inductive bias. In particular, combining both physics-informed 
terms yielded the fastest convergence, lowest error, and strongest predictive 
stability compared with purely data-driven approaches. The trained NASP--PINN 
accurately reproduced long-term SISR dynamics and captured the dependence of the 
$\mathrm{CV}$ on noise intensity, excitability, and timescale separation, demonstrating its 
ability to faithfully encode the underlying stochastic physics.

Although in this work the governing SDEs of the FHN neuron are fully known and therefore directly incorporated as physical priors (phy1 and phy2), the framework extends naturally to scenarios where the equations are partially known or only qualitatively specified. In such cases, data-only terms can be used to fit the unknown components, while the available physics---such as conservation laws, timescale separation, escape-rate scaling, or qualitative dynamical structure---can still be incorporated as weak constraints or surrogate regularizers. Thus, the proposed NASP–PINN model is not restricted to analytically tractable systems, but can act as a hybrid data–physics learner for experimental recordings where only partial mechanistic knowledge is available.

Figure~\ref{fig:10} confirmed that the PINN's predictions quantitatively replicate the noise--coherence profiles obtained from direct stochastic simulations, including the shifts of $\mathrm{CV}$ minima with increasing~$a$ and~$\varepsilon$. This agreement highlights the model's capacity to generalize beyond the training data and to serve as an efficient surrogate for simulating slow--fast stochastic systems, where conventional numerical approaches can be computationally prohibitive.  

In addition, this physics-informed neural framework aligns with current trends in scientific machine learning, where neural ordinary differential equations~(neural ODEs) \cite{worsham2025guide}, reservoir computing \cite{kobiolka2025reduced}, and physics-informed neural networks serve as data-efficient surrogates for complex dynamical systems. Beyond predictive accuracy, the proposed NASP-PINN framework provides both a computational solver and an interpretive model, underscoring its dual role as a practical simulator and a theoretical tool for analyzing noise-driven dynamics. It thus reaffirms its value as a powerful and general approach for studying complex stochastic phenomena in which purely analytical or numerical methods may be insufficient.

Future work will explore the unknown (or partially known) model setting, in which neural
networks learn dynamics using drifts or noise terms identified through sparse regression
methods (\textit{e.g.,} SINDy), \marius{and} physics-based SISR constraints regularize the solution space. This direction
will enable application of the framework to biological neuron data, where no (or only partial) mathematical models are available.

\section*{Acknowledgements} This work was funded by the Department of Data Science (DDS), Friedrich-Alexander-Universit\"at Erlan\-gen-Nürnberg, Germany, and the Deutsche Forschungsgemeinschaft (DFG, German Research Foundation) via the grant YA 764/1-1 to M.E.Y—Project No. 456989199.

\section*{Appendix}\label{appendix} 
\begin{table}[h]\label{tab:2}
\centering
\caption{Definition of notations used in Algorithm \ref{alg:1}}
\begin{tabular}{ll}
\toprule
Symbol & Definition \\
\midrule
$v_t$ & Fast membrane potential variable \\
$w_t$ & Slow recovery current variable \\
$\eta_t$ & One-dimensional Brownian motion on $v_t$ variable\\
$\sigma$ & Noise amplitude \\
NASP & One-step, noise-aware neural integrator
$(v_t, w_t, \sigma \eta_t) \mapsto (\hat{v}_{t+\Delta t}, \hat{w}_{t+\Delta t})$ \\
$e_{\max}$ & Maximum number of training epochs \\
$\Delta t$ & Time step \\
$\lambda_{\mathrm{data}},\lambda_{\mathrm{ic}},\lambda_{\mathrm{phy1}},\lambda_{\mathrm{phy2}}$ & Loss weights \\
$B$ & Mini-batch size (forms a $B\times T$ window) \\
$T$ & Window length (sequence length in the $B\times T$ window) \\
$\theta$ & Model parameters (weights and biases of the MLP) \\
$\alpha$ & Learning rate (used in $\text{Adam}(\theta,\alpha)$) \\
$O$ & Optimizer \\
$|v_t|$ & Dataset length used when sampling indices \\
$I$ & Mini-batch indices \\
$\hat w_{\ell},\hat w_{r}$ & Left and right predicted escape points of $\hat{w}_{t+\Delta t}$ \\
$\Delta U_{\ell}(\hat w_{\ell},a),\ \Delta U_{r}(\hat w_{r},a)$ & Left and right potential barriers \\
\bottomrule
\end{tabular}
\end{table}

\begin{algorithm}[h]
  \SetKwComment{Comment}{// }{}
  \KwInput{$(v_t,\,w_t,\,\sigma\eta_t) (\text{inputs to NASP}),\;e_{\max},\,\alpha,\,\Delta t,\,\lambda_{\mathrm{data}},\,\lambda_{\mathrm{ic}},\,\lambda_{\mathrm{phy1}},\,\lambda_{\mathrm{phy2}},\,B,\,T$}
  \KwOutput{$\theta$}

  \For{$e \gets 1$ \KwTo $e_{\max}$}{
    \If{$e = 1$}{
      $\theta \leftarrow \text{NASP}(3\!\to\!128\!\to\!128\!\to\!128\!\to\!2)$\Comment*[r]{NASP in MLP (Sec.~\ref{sec:mlp})}%
      $O \leftarrow \text{Adam}(\theta,\alpha)$\Comment*[r]{Optimizer}%
    }

     $I \leftarrow \text{RandIdx}(|v_t|,B)$ \Comment*[r]{mini-batch indices}%
     $(v_{1:T},w_{1:T},\sigma\eta_{1:T}) \leftarrow (v_t[I,1:T],w_t[I,1:T],\sigma\eta_t[I,1:T])$ \Comment*[r]{take $B\times T$ window}%
     $(\hat v_{2:T},\hat w_{2:T}) \leftarrow \theta(v_{1:T-1},w_{1:T-1},\sigma\eta_{2:T})$ \Comment*[r]{predict next step}%

     $\Lb_{\mathrm{data}} \leftarrow \mathrm{MSE}\!\big[\hat v_{t+\Delta t},\, v_{t+\Delta t}\big] + \mathrm{MSE}\!\big[\hat w_{t+\Delta t},\, w_{t+\Delta t}\big]$\Comment*[r]{Data loss}%

     $(\hat v_{0},\hat w_{0}) \leftarrow \theta(v_0,w_0,\sigma\eta_0)$ \;
     $\Lb_{\mathrm{ic}} \leftarrow \mathrm{MSE}\!\big[\hat v_{0},v_{0}\big] + \mathrm{MSE}\!\big[\hat w_{0},w_{0}\big]$\Comment*[r]{Ic loss}%

     $(\frac{d\hat v}{dt},\frac{d\hat w}{dt})
   \leftarrow \text{AutoDiff}\big(\theta;\, v_t, w_t, \sigma\eta_{t+\Delta t}\big)$\tcp*[r]{predicted time-derivatives via\\Autodiff}
    
     $\Lb_{\mathrm{phy1}} \leftarrow \mathrm{MSE}\!\big[\frac{d\hat v}{dt},\,f(v_t,w_t) + \sigma\eta_{t+\Delta t}\big] + \mathrm{MSE}\!\big[\frac{d\hat w}{dt},\,g(v_t,w_t)\big]$\Comment*[r]{phy1 loss, where\\$f(v,w)=v(a-v)(v-1)-w$ and $g(v,w)=\varepsilon(bv-cw)$}%

     $\hat w_{0:T} \leftarrow \text{Predicted time series}(\theta, v_0, w_0, \sigma\eta_0, \Delta t)$ \;
     $(\hat w_{\ell},\hat w_{r}) \leftarrow \text{Extrema}(\hat w_{0:T})$ \;
     $\Delta U_{\ell} \leftarrow \Delta U_{\ell}(\hat w_{\ell},a)$ \; $\Delta U_{r} \leftarrow \Delta U_{r}(\hat w_{r},a)$ \;
     $\tfrac12\,\sigma^{2}\log(\varepsilon^{-1}) \leftarrow \text{Deterministic timescale}$ \;
    
     $\Lb_{\mathrm{phy2}} \leftarrow \mathrm{MSE}\!\big[\tfrac12\,\sigma^{2}\log(\varepsilon^{-1}),\,\Delta U_{\ell}\big]+\mathrm{MSE}\big[\tfrac12\,\sigma^{2}\log(\varepsilon^{-1}),\,\Delta U_{r}\big]$ \Comment*[r]{\\phy2 loss}%

     $\Lb \leftarrow \lambda_{\mathrm{data}}\Lb_{\mathrm{data}} + \lambda_{\mathrm{ic}}\Lb_{\mathrm{ic}} + \lambda_{\mathrm{phy1}}\Lb_{\mathrm{phy1}} + \lambda_{\mathrm{phy2}}\Lb_{\mathrm{phy2}}$\Comment*[r]{Total loss}%
     $O.\text{zero\_grad}()$\Comment*[r]{clear old gradients}%
     $\nabla_{\!\theta}\Lb$\Comment*[r]{backpropagation to get gradients}%
     $O.\text{step}()$\Comment*[r]{update weights}%
  }
  \KwRet{$\theta$}\Comment*[r]{return trained parameters}%
  \caption{NASP-PINN algorithm for predicting SISR}
  \label{alg:1}
\end{algorithm}

\section*{Data availability}
The codes developed and used to generate the data supporting the findings
of this study are publicly accessible \cite{code}.

\newpage
\providecommand{\noopsort}[1]{}\providecommand{\singleletter}[1]{#1}

\end{document}